\documentclass[lettersize,journal]{IEEEtran}
\usepackage{amsmath,amsfonts}
\usepackage{array}
\usepackage[caption=false,font=normalsize,labelfont=sf,textfont=sf]{subfig}
\usepackage{textcomp}
\usepackage{stfloats}
\usepackage{url}
\usepackage{verbatim}
\usepackage{graphicx}
\usepackage{cite}
\usepackage{graphicx}
\usepackage{amsmath}
\usepackage{amssymb}
\usepackage{booktabs}
\usepackage{amsthm,amsmath,amssymb}
\usepackage{mathrsfs}

\usepackage{amsmath}
\DeclareFontFamily{U}{mathc}{}
\DeclareFontShape{U}{mathc}{m}{it}%
{<->s*[1.03] mathc10}{}

\DeclareMathAlphabet{\mathscr}{U}{mathc}{m}{it}

\usepackage{array, tabularx, boldline}
\usepackage{graphicx}
\usepackage{cellspace}
\usepackage{color}
\usepackage[dvipsnames, table]{xcolor}

\usepackage{times}
\usepackage{epsfig}
\usepackage{graphicx}
\usepackage{amsmath}
\usepackage{amssymb}
\usepackage{amsthm}
\usepackage{algpseudocode}
\usepackage[ruled]{algorithm2e}

\newcommand{\etal}{\textit{et al}.}
\newcommand{\ie}{\textit{i}.\textit{e}.}
\newcommand{\eg}{\textit{e}.\textit{g}.}
\newcommand{\etc}{\textit{etc}.}

\usepackage{graphicx}
\usepackage{amsmath}
\usepackage{amssymb}
\usepackage{booktabs}
\usepackage{float}
\usepackage{times}
\usepackage{pifont}
\usepackage{enumitem}
\usepackage{multirow}

\newcommand{\framework}[1]{CLIP-VG}
\definecolor{myblue}{rgb}{0.27, 0.80, 1.0}
\definecolor{mygreen}{rgb}{0.6, 1.0, 0.6}
\definecolor{myred}{rgb}{1.0, 0.2, 0.2}

\usepackage[pagebackref=false,breaklinks=true,letterpaper=true,colorlinks,bookmarks=true]{hyperref}

\hypersetup{
    colorlinks=true,
    linkcolor=black,
    filecolor=black,      
    urlcolor=cyan, % this can be comment to change the color to pink
    citecolor=blue,
}

\newlength\secmargin
\newlength\subsecmargin
\newlength\paramargin
\newlength\figmargin
\newlength\eqmargin
\usepackage[capitalize]{cleveref}
\crefname{section}{Sec.}{Secs.}
\Crefname{section}{Section}{Sections}
\Crefname{table}{Table}{Tables}
\crefname{table}{Tab.}{Tabs.}
\crefname{algorithm}{Algo.}{Algos.}

\usepackage{makecell}
\usepackage{colortbl}
\usepackage{booktabs}
\usepackage{multirow}

\usepackage{bm}

\usepackage{orcidlink} %调包

\begin{document}
%%%%%%%%%%%%%%%%%%%%%%%%%%%%%%%%%%%%%%%%%%%%%%%%%%%%%%%%%%%%%%%%%%
%----------------------
%----------------------
%----------------------
\title{CLIP-VG: Self-paced Curriculum Adapting of CLIP for Visual Grounding}

\author{{Linhui~Xiao$^{\orcidlink{0000-0003-2592-5264}}$, Xiaoshan~Yang$^{\orcidlink{0000-0001-5453-9755}}$, Fang~Peng$^{\orcidlink{0000-0002-3948-7413}}$, Ming~Yan$^{\orcidlink{0000-0003-4959-8878}}$, Yaowei~Wang$^{\orcidlink{0000-0002-6110-4036}}$, \\ and~Changsheng~Xu$^{\orcidlink{0000-0001-8343-9665}}$, \IEEEmembership{Fellow,~IEEE}}
% <-this % stops a space
\thanks{Linhui Xiao, Xiaoshan Yang, Fang Peng and Changsheng Xu are with the State Key Laboratory of Multimodal Artificial Intelligence Systems (MAIS), Institute of Automation, Chinese Academy of Sciences (CASIA), Beijing 100190, China, also with the Peng Cheng Laboratory (PCL), Shenzhen 518066, China, and also with the School of Artificial Intelligence, University of Chinese Academy of Sciences (UCAS), Beijing 100049, China (e-mail: xiaolinhui16@mails.ucas.ac.cn, xiaoshan.yang@nlpr.ia.ac.cn, pengfang21@mails.ucas.ac.cn, csxu@nlpr.ia.ac.cn).} % <-this % stops a space
\thanks{Ming Yan is with the DAMO Academy, Alibaba Group, Hangzhou 311121, China (e-mail: ym119608@alibaba-inc.com). Yaowei Wang is with the Peng Cheng Laboratory, Shenzhen 518066, China (e-mail: wangyw@pcl.ac.cn).} % <-this % stops a space
\thanks{Changsheng Xu is the corresponding author.} % <-this % stops a space
\thanks{This work is supported by the National Natural Science Foundation of China (No. 62036012, 62322212, 62072455), and also supported by Peng Cheng Laboratory Research Project No. PCL2023AS6-1.}
\thanks{Digital Object Identifier \url{https://doi.org/10.1109/TMM.2023.3321501}}
}

% \markboth{Journal of \LaTeX\ Class Files,~Vol.~14, No.~15, May~2023}%
\markboth{IEEE Transaction on Multimedia,~Vol.26, 2023}%
{Shell \MakeLowercase{\textit{et al.}}: A Sample Article Using IEEEtran.cls for IEEE Journals}

\IEEEpubid{1520-9210~\copyright~2023 IEEE. Personal use is permitted, but republication/redistribution requires IEEE permission.}

\maketitle

\begin{abstract}

Visual Grounding (VG) is a crucial topic in the field of vision and language, which involves locating a specific region described by expressions within an image. To reduce the reliance on manually labeled data, unsupervised visual grounding have been developed to locate regions using pseudo-labels. However, the performance of existing unsupervised methods is highly dependent on the quality of pseudo-labels and these methods always encounter issues with limited diversity. In order to utilize vision and language pre-trained models to address the grounding problem, and reasonably take advantage of pseudo-labels, we propose CLIP-VG, a novel method that can conduct self-paced curriculum adapting of CLIP with pseudo-language labels. We propose a simple yet efficient end-to-end network architecture to realize the transfer of CLIP to the visual grounding. Based on the CLIP-based architecture, we further propose single-source and multi-source curriculum adapting algorithms, which can progressively find more reliable pseudo-labels to learn an optimal model, thereby achieving a balance between reliability and diversity for the pseudo-language labels. Our method outperforms the current state-of-the-art unsupervised method by a significant margin on RefCOCO/+/g datasets in both single-source and multi-source scenarios, with improvements ranging from 6.78$\%$ to 10.67$\%$ and 11.39$\%$ to 14.87$\%$, respectively. The results even outperform existing weakly supervised methods. Furthermore, our method is also competitive in fully supervised setting. The code and models are available at \url{https://github.com/linhuixiao/CLIP-VG}.

\end{abstract}


%%%%%%%%%%%%%%%%%%%%%%%%%%%%%%%%%%%%%%%%%%%%%%%%%%%%%%%%%%%%%%%%%%
%----------------------
%----------------------
%----------------------

\section{Introduction}
\label{sec:intro}

\begin{figure}[t]
 \centering
   \includegraphics[width=0.95\linewidth]{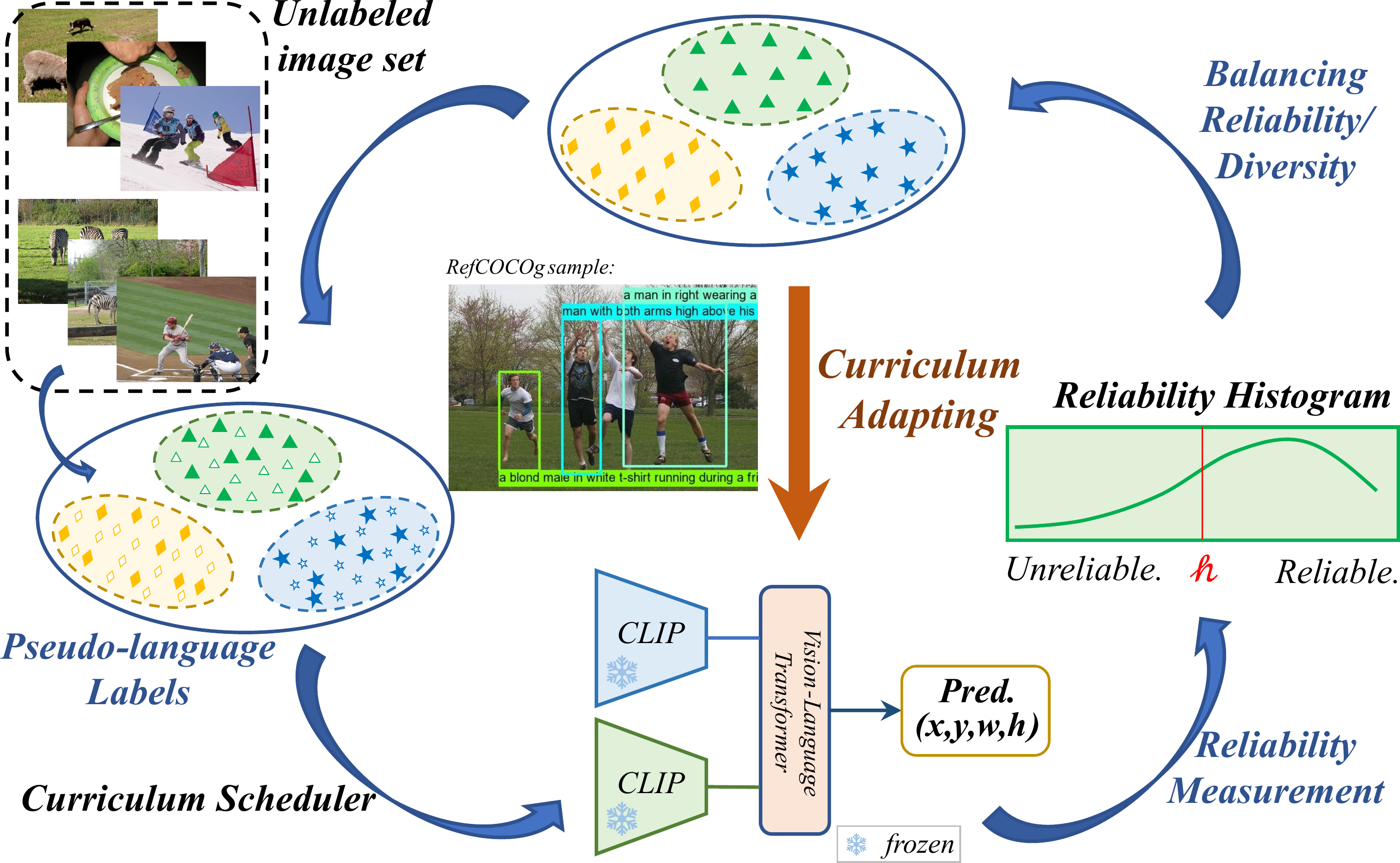}
   \vspace{-5pt}
   \caption{Main idea of our proposed CLIP-VG, which adapts CLIP with pseudo-language labels in a self-paced curriculum adapting paradigm to realize the transfer learning in visual grounding.}
    \vspace{-10pt}
   \label{fig:fig1}
\end{figure}

\IEEEPARstart{V}{isual} Grounding (VG) \cite{qiao2020referring,mao2016generation,yu2016modeling,hu2016natural,deng2021transvg}, also known as Referring Expression Comprehension (REC) or Phrase Grounding (PG), refers to locating the \texttt{bounding box} (\ie, \texttt{bbox}) region described by a textual \texttt{expression} in a specific \texttt{image}, which has become one of the critical technologies in various Vision-Language (V-L) fields, such as visual question answering \cite{antol2015vqa} and visual language navigation \cite{anderson2018vision}. Due to its cross-modal properties, the grounding model is required to comprehend the semantics of both language expressions and images, which has long been a challenging task. Considering its task complexity, most existing methods focus on fully supervised setting (\textit{i.e., using manual triplet-paired data as supervised signal}) \cite{chen2018real,liao2020real,hong2019learning,hu2017modeling,liu2019learning,deng2021transvg}. Nevertheless, high-quality annotation is strictly required for supervised grounding. Specifically, the \texttt{expression} needs to be paired with \texttt{bbox}, unique in referring, and rich in semantics. To reduce the reliance on labor-intensive labeled data, weakly supervised (\textit{i.e., only given image and query pairs, no paired bbox}) \cite{sun2021cycle,wang2021weakly,chen2018knowledge,datta2019align2ground,gupta2020contrastive,liu2021relation,wang2021improving} and unsupervised grounding (\textit{i.e., locating image regions without using any task-related annotations}) \cite{yeh2018unsupervised, wang2019phrase, shi2022unpaired, jiang2022pseudo} have recently gained increasing attention.

\IEEEpubidadjcol

Existing unsupervised visual grounding methods~\cite{yeh2018unsupervised, wang2019phrase, shi2022unpaired} mainly realized referring grounding with unpaired data by exploiting pre-trained detectors and an additional large-scale corpus. The state-of-the-art (SOTA) unsupervised method \cite{jiang2022pseudo} proposes using manually designed templates and spatial relationship prior knowledge to match the results obtained by the object and attribute detectors, along with the corresponding object \texttt{bbox}. This generates \texttt{expression} and \texttt{bbox} pseudo pairs, which are used as pseudo-labels to learn the grounding model in a supervised manner. However, the effectiveness of the pseudo annotations in these existing methods heavily relies on the object or attribute detectors that are always pre-trained on a specific dataset. This can limit the diversity of the language taxonomy and match patterns, as well as the contextual semantics richness, ultimately harming the model generalization ability.

In the past couple of years, the Vision-Language Pre-trained (VLP) foundation models (\textit{e.g.}, CLIP \cite{radford2021learning}) have achieved impressive results on many downstream tasks through adapting or prompting paradigm with a few task-related data. The main advantage of these foundation models is that they can learn general knowledge from the readily available web data and various downstream task data (\eg, BeiT3\cite{wang2022image}) with self-supervised constraints. This inspires us to consider transferring the VLP models (\textit{i.e.}, CLIP is used in this work) to solve the downstream grounding task in an unsupervised manner. This is a challenging task due to the lack of task-related labeled data. A straightforward solution is to leverage the pseudo annotations generated in previous unsupervised grounding methods to fine-tune the pre-trained model. However, this will impact the generalization ability of the pre-trained model due to the gap between the pseudo annotations and the ground-truth task-specific annotations.

In this paper, we propose \textbf{CLIP-VG}, as shown in \cref{fig:fig1}, a novel method that can conduct self-paced curriculum adapting of CLIP via exploiting pseudo-language labels to address the visual grounding problem. Firstly, we propose a simple yet efficient end-to-end pure-Transformer encoder-only network architecture. It only requires adapting a few parameters and costing minimal training resources to realize the transfer of CLIP to visual grounding. Secondly, to achieve a more stable adaption of the CLIP-based network architecture by finding reliable pseudo-labels, we propose a scheme for evaluating instance-level quality and a progressive adapting algorithm based on Self-Paced Curriculum Learning (SPL), namely Reliability Measurement (\cref{3.3_reliability}) and Single-source Self-paced Adapting (SSA) algorithm (\cref{3.4SSA}). The instance-level Reliability is calculated as the likelihood of being correctly predicted by a measurer model that is learned with a specific label source. Specifically, we learn a preliminary grounding model as Reliability Measurer with CLIP as the backbone for the pseudo-labels and then score the samples' reliability to construct a Reliability Histogram (RH). Next, according to the constructed RH, the SSA algorithm is executed in a self-paced manner, progressively sampling more reliable pseudo-labels to improve the grounding performance. To efficiently select a subset of pseudo-paired data, we design a greedy sample selection strategy based on the modified binary search to achieve an optimal balance between reliability and diversity.

One major advantage of the proposed CLIP-VG is that its progressive adapting framework is not dependent on the specific form or quality of the pseudo-labels. Therefore, the CLIP-VG can be flexibly extended to access multiple sources of pseudo-labels. In the multi-source scenario, we first independently learn a preliminary source-specific grounding model for each pseudo-label source. Then, we propose the source-level complexity metric. Specifically, in different steps of the SPL, we gradually select the pseudo-label source from simple to complex according to \textit{the average number of entities per expression}. Based on SSA, we further propose Source-specific Reliability (SR) and Cross-source Reliability (CR), as well as a Multi-source Self-paced Adapting (MSA) algorithm (\cref{3.5MSA}). The source-specific reliability is calculated as the likelihood of being correctly predicted by the grounding model learned with the current label source. In contrast, cross-source reliability is calculated as the likelihood of being correctly predicted by grounding models learned with other label sources. Thus, the whole method can progressively utilize pseudo-labels to learn the grounding model in an easy-to-hard curriculum paradigm, which maximizes the exploitation of different source pseudo-labels and ensures the generalization of the foundation model.

On the five mainstream benchmarks, RefCOCO/+/g \cite{yu2016modeling, mao2016generation}, ReferitGame \cite{kazemzadeh2014referitgame} and Flickr30K Entities \cite{plummer2015flickr30k}, our model outperforms the SOTA unsupervised grounding method Pseudo-Q \cite{jiang2022pseudo} in both single-source and multi-source scenarios with a significant margin, \textit{i.e.}, \textbf{6.78$\%$$\sim$10.67$\%$} and \textbf{11.39$\%$$\sim$ 14.87$\%$}, respectively. The performance gains brought by the proposed SSA and MSA algorithms are \textbf{3+$\%$}. Furthermore, our approach even outperforms existing weakly supervised methods. In comparison with the fully supervised SOTA model, QRNet\cite{ye2022shifting}, we achieve comparable results with \textbf{only 7.7$\%$} of its updated parameters, while obtaining significant speedups in both training and inference, up to \textbf{26.84$\times$} and \textbf{7.41$\times$}, respectively. Compared to the reported results \cite{ho2023yoro}, our model also achieves the SOTA in terms of both speed and energy efficiency.

In summary, the contributions of this paper are four-fold:

\begin{itemize}\setlength{\itemsep}{0pt}
% \vspace{-6pt}
\setlength{\parsep}{0pt}
\setlength{\parskip}{0pt}
    \item
    As far as we know, we are the first to adapt CLIP to realize unsupervised visual grounding. Our method can transfer the cross-modal learning ability of CLIP to visual grounding with only a small training cost.
    \item
    We are the first to introduce self-paced curriculum learning in unsupervised visual grounding. Our proposed reliability measurement and single-source self-paced adapting can progressively enhance the CLIP-based visual grounding model by utilizing pseudo-labels in an easy-to-hard learning paradigm.
    \item
    We first propose the multi-source self-paced adapting algorithm to extend our method for accessing multiple sources of pseudo-labels, which can flexibly improve the diversity of language taxonomy.
    \item
    We conduct extensive experiments to evaluate the effectiveness of our approach. Results show that our method obtains significant improvements in unsupervised setting and is also competitive in fully supervised setting. 
\end{itemize}

%%%%%%%%%%%%%%%%%%%%%%%%%%%%%%%%%%%%%%%%%%%%%%%%%%%%%%%%%%%%%%%%%%%%%%%%%%%
%----------------------
%----------------------
%----------------------
\section{Related Work}
\label{sec:RelatedWork}

\subsection{Visual Grounding}
% \vspace{-0pt}

Visual Grounding (VG) involves both visual and linguistic modalities. With the advancement of Transformer \cite{vaswani2017attention} and ViT \cite{dosovitskiy2020image, carion2020end}, the technical route of VG is changing from traditional CNN-based \cite{chen2018real,liao2020real,sadhu2019zero,yang2020improving,hu2017modeling,liu2019learning,yu2018mattnet,wang2019neighbourhood} to the Transformer-based approach \cite{jiang2022pseudo, deng2021transvg, ye2022shifting, kamath2021mdetr}. Recent VG methods can be summarized into five categories: fully supervised \cite{chen2018real,liao2020real,hong2019learning,hu2017modeling,liu2019learning,deng2021transvg}, weakly supervised \cite{xiao2017weakly,datta2019align2ground,gupta2020contrastive,liu2021relation,wang2021improving,sun2021discriminative}, semi-supervised \cite{zhu2021utilizing, chou2022semi}, unsupervised \cite{yeh2018unsupervised, wang2019phrase, jiang2022pseudo} and zero-shot \cite{sadhu2019zero, subramanian2022reclip, li2022adapting}. 
Without using any image sample and labeled data, zero-shot work, \eg, ReCLIP\cite{subramanian2022reclip} and adapting-CLIP\cite{li2022adapting}, utilize pre-trained detectors to extract proposals, thus achieving training-free grounding capabilities. Previous unsupervised methods\cite{yeh2018unsupervised, wang2019phrase, shi2022unpaired} attempt to solve this problem by using unpaired image-query based on pre-trained detectors and large-scale corpus. However, image-query and query-box double pairing under this approach will meet the challenge. Pseudo-Q \cite{jiang2022pseudo} proposes to generate template pseudo-labels based on the detectors, which directly eliminates the error caused by double pairing. Different from Pseudo-Q, we propose self-paced adapting algorithms to find a balance between reliability and diversity for any pseudo-labels in visual grounding.

\vspace{-5pt}
\subsection{Vision-Language Pre-trained Models}
% \vspace{-2pt}

Transformer-based cross-modal Vision-Language Pre-trained (VLP) models emerge in an endless stream. A series of work, \eg, CLIP\cite{radford2021learning}, M6\cite{lin2021m6}, ALBEF\cite{li2021align}, OFA\cite{wang2022unifying}, BeiT3\cite{wang2022image} \textit{etc.} are trained on massive data by leveraging contrastive learning and mask modeling, constantly refreshing the SOTA in various tasks \cite{radford2021learning, li2021align, wang2022unifying, peng2023sgva}. In order to leverage the generalization ability of the VLP models, we build our model on CLIP while considering its scalability for achieving cross-modal grounding.

\vspace{-4pt}
\subsection{Curriculum Learning}
% \vspace{-2pt}

Curriculum Learning (CL), as proposed by Bengio \etal \cite{bengio2009curriculum}, is a training strategy that trains machine learning models from easy to hard, which mimics the process of human learning curricula. The strategy of CL usually performs its power in improving the generalization and denoising in various computer vision (CV) and natural language processing (NLP) tasks \cite{shu2019transferable,soviany2022curriculum,wang2021survey}. There are many CL-based unsupervised or semi-supervised works that focus on pseudo-labeling\cite{choi2019pseudo,cascante2021curriculum}. Most of them are in NLP \cite{zhang2021review, tay2019simple}, classification \cite{gong2016multi} and detection \cite{zhao2022exploiting} tasks, where the pseudo-labels are relatively simple \cite{wang2021survey}. However, there are few CL works that focus on more complex cross-modal tasks (\eg, VQA, VLN, VG) due to the difficulty in evaluating data and models with diverse modalities and task targets \cite{soviany2022curriculum,wang2021survey}. Self-Paced Curriculum Learning (SPL) \cite{kumar2010self} is semi-automatic CL with a dynamic curriculum, which takes the training loss of the current model as the criteria and realizes the automation of difficulty measurement \cite{wang2021survey}. Our work is designed based on the SPL paradigm.

%%%%%%%%%%%%%%%%%%%%%%%%%%%%%%%%%%%%%%%%%%%%%%%%%%%%%%%%%%%%%%%%%%%%%%%%%
%------------------------------------------------------------------------
%------------------------------------------------------------------------
%------------------------------------------------------------------------
%------------------------------------------------------------------------
\vspace{-3pt}
\section{Method}
\vspace{2pt}

We propose \textbf{CLIP-VG}, a novel method that can conduct self-paced curriculum adapting of CLIP via exploiting pseudo-language labels to address the visual grounding problem. Our approach mainly includes (1) a simple yet efficient CLIP-based pure-Transformer visual grounding model, (2) a sample reliability evaluation scheme, (3) a self-paced adapting algorithm in a single-source scenario, and (4) a further extended multi-source self-paced adapting algorithm. In this section, we will first provide the Task Definition (\cref{3.1_task_definition}) and then present our method, which includes Network Architecture (\cref{3.2framework}), Reliability Measurement (\cref{3.3_reliability}), Single-source Self-paced Adapting (SSA) (\cref{3.4SSA}), and Multi-source Self-paced Adapting (MSA) (\cref{3.5MSA}).

\vspace{-2pt}
\subsection{Task Definition}
\label{3.1_task_definition}
\vspace{-2pt}

Our approach follows the setting of the previous state-of-the-art unsupervised method Pseudo-Q \cite{jiang2022pseudo}, \ie, without using any task-related annotation during training.

Define $\mathcal{I}$ as the unlabeled image dataset. By utilizing the generated pseudo-labels, we construct a single-source \textbf{pseudo triplet-paired set}, denoted as $\mathcal{D}^{s}=\{\mathcal{S}\}$, where $\mathcal{S}=\left(\mathcal{I},\ \mathcal{E},\ \mathcal{B}\right)$, and $\mathcal{E}$ represents the set of pseudo expressions, $\mathcal{B}$ represents the set of pseudo bounding boxes. The test dataset is defined as $\mathcal{D}^{t}=\left(\mathcal{I}_t,\ \mathcal{E}_t,\ \mathcal{B}_t\right)$. We aim to learn a model $\mathcal{F}_\theta:\ (\mathcal{I},\ \mathcal{E})\rightarrow\mathcal{B}$ based on $\mathcal{D}^{s}$ so that it can generalize well on the test data $\mathcal{D}^{t}$:
% \vspace{-5pt}
\begin{equation}
% \vspace{-4pt}
\mathcal{F}_{\theta}^{*}=\arg \min _{\mathcal{F}_{\theta}} \ell\big(\mathcal{F}_{\theta}(\mathcal{I},\mathcal{E}),\mathcal{B}\big),
\vspace{-0pt}
\label{eq:model_s}
\end{equation}

\noindent where $\ell$ represent loss function, which measures the distance between the predicted bbox and pseudo bbox by leveraging smooth L1 loss \cite{girshick2015fast} and Giou loss \cite{rezatofighi2019generalized} with coefficient $\lambda$:
% \vspace{-4pt}
\begin{equation}
% \small
\ell=\mathcal{L}_{\text {smooth-l1 }}\big(\mathcal{F}_{\theta}(\mathcal{I},\mathcal{E}),\mathcal{B}\big)+\lambda\cdot\mathcal{L}_{\text{giou}}\big(\mathcal{F}_{\theta}(\mathcal{I},\mathcal{E}),\mathcal{B}\big).
% \vspace{-5pt}
\label{eq:loss1}
\end{equation}

In this work, we also consider the problem of multi-source pseudo-labels. Assuming that there are multiple sources of triplet-paired pseudo-labels generated by different ways, denote as $\mathcal{D}^{s}=\{\mathcal{S}_i\}_{i=1}^n$, where $\mathcal{S}_i=\left(\mathcal{I},\ \mathcal{E}_i,\ \mathcal{B}_i\right)$, $\mathcal{E}_i$ represents the set of pseudo expressions from the $i$-th source and $\mathcal{B}_i$ represents the set of bbox from $i$-th source. Then, the aim of the model becomes:
% \vspace{-2pt}
\begin{equation}
\vspace{-8pt}
\mathcal{F}_{\theta}^{*}=\arg \min _{\mathcal{F}_{\theta}} \Sigma_{i=1}^n\ell\big(\mathcal{F}_{\theta}(\mathcal{I},\mathcal{E}_i),\mathcal{B}_i\big).
\vspace{-2pt}
\label{eq:model_m}
\end{equation}

\begin{figure}[t]
\centering
   \includegraphics[width=1.0\linewidth]{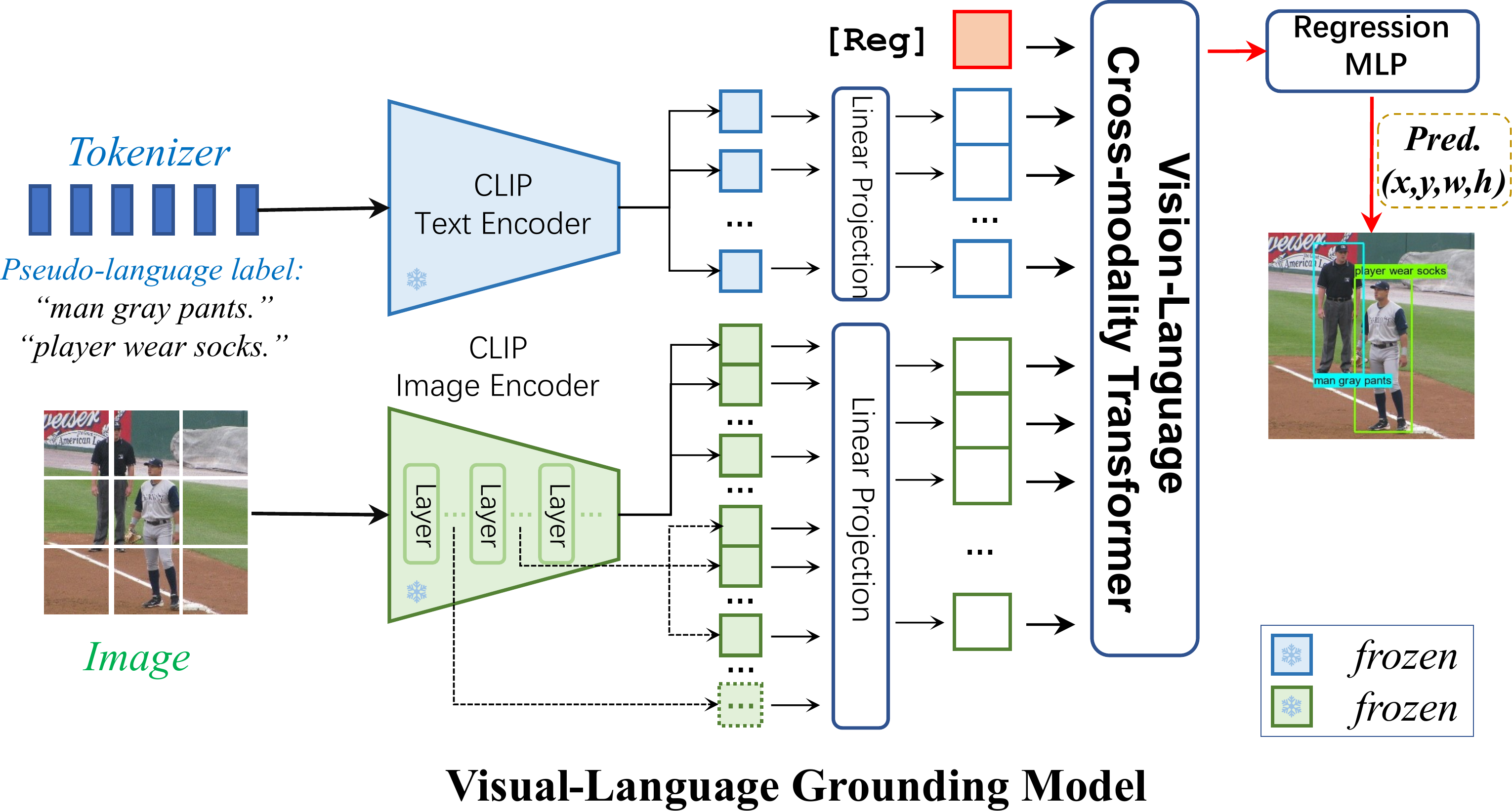}
   \vspace{-15pt}
   \caption{Our CLIP-VG model architecture (\cref{3.2framework}) serves as a vision-language grounding model to realize the self-paced curriculum adapting of CLIP.}
   \vspace{-10pt}
   \label{fig:fig3}
\end{figure}

\begin{figure*}[t]
\centering
   % \vspace{\figmargin}
   \includegraphics[width=1.0\linewidth]{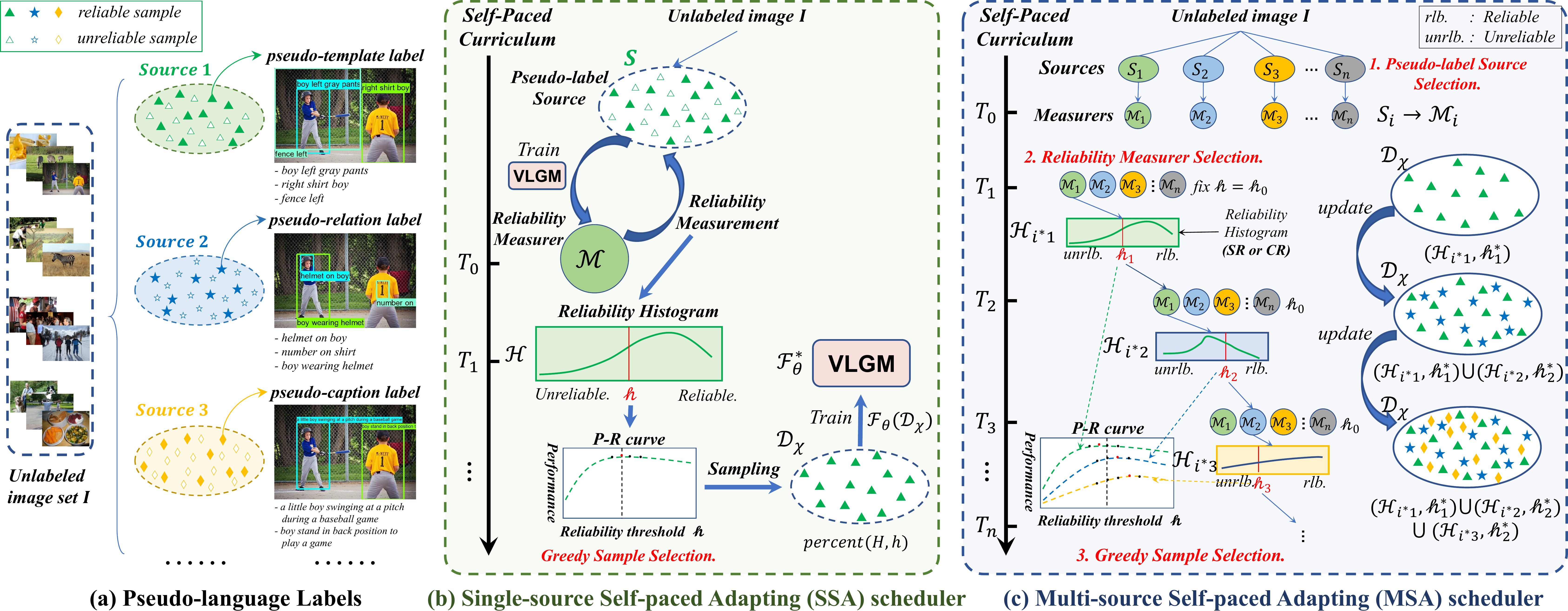}
   \vspace{-15pt}
   \caption{Self-paced curriculum adapting of CLIP by exploiting pseudo-language labels to realize the unsupervised visual grounding. (a) Examples of pseudo-language labels (The sources of different pseudo-language labels are described in \cref{4.1detail}, better view in zoom-in). (b) Single-source Self-paced Adapting (SSA) utilizes the vision-language grounding model (VLGM) to exploit the pseudo-template labels for reliability measurement and greedy sample selection to achieve a more stable adaption of the CLIP by finding reliable pseudo-labels. (c) Multi-source Self-paced Adapting (MSA) further proposes source-specific reliability (SR) and cross-source reliability (CR) based on SSA. It sequentially conducts pseudo-label sources selection, reliability measurer selection, and greedy sample selection to achieve an optimal balance between reliability and diversity.}
   \vspace{\figmargin}
\label{fig:fig4}
\vspace{-5pt}
\end{figure*}

\subsection{Network Architecture}
% \vspace{-3pt}
\label{3.2framework}

Since CLIP is pre-trained under the image-level vision-language contrastive constraints, it lacks region-level grounding capabilities. To enable the transfer learning of CLIP on the grounding task while adapting only a few parameters, we only connected a 6-layer vision and language cross-modal vanilla Transformer encoder \cite{carion2020end}. The illustration of the CLIP-VG model can be seen in \cref{fig:fig3}. Our model incorporates two CLIP encoders and a Transformer encoder. To better utilize scale information, we propose extracting multi-layer visual intermediate features $\{\bm{f}_v^i\}_{i=1}^n \in \mathbb{R}^{B\times N_v \times H_{clip}}$ from the CLIP image encoder layers and concatenating them along the hidden dimension. Then, we project them into a visual embedding $\bm{p}_v \in \mathbb{R}^{B\times N_v \times H_{cross}}$ with the same hidden dimension $H_{cross}$ as that of the cross-modality Transformer to perceive multi-layer visual representations:
\begin{equation}
{\bm{p}_v = {\rm{concat}} [\bm{f}_v^1, \bm{f}_v^2, \cdots, \bm{f}_v^n] \times W_v},
\label{eq:visu_proj}
\end{equation}
where $n$ represents the number of extracted layers, $B$ represents the batch size, $N_v$ represents the token length of CLIP visual features, $H_{clip}$ represents the hidden dimension size of CLIP, and $W_v\in\mathbb{R}^{(n \cdot H_{clip}) \times H_{cross}}$ represents the weight for visual projection. For language modality, we only project the last layer feature $\bm{f}_l^{last} \in \mathbb{R}^{B\times N_l \times H_{clip}}$ of the CLIP text encoder into a language embedding $\bm{p}_l \in \mathbb{R}^{B\times N_l \times H_{cross}}$ with a language projection weight $W_l\in\mathbb{R}^{H_{clip} \times H_{cross}}$:
\begin{equation}
{\bm{p}_l = \bm{f}_l^{last} \times W_l}.
\label{eq:visu_proj}
\end{equation}
The token order input to the cross-modal Transformer is as follows:
% \vspace{-5pt}
\begin{equation}
\small
\mathcal{X} = [\ \mathscr{p}_r\ ,  \overbrace{p_l^1,\ p_l^2,\ \cdots,\ p_l^{N_l}}^{\text{CLIP language tokens}~\bm{p}_l},\ \underbrace{cls^1,\ p_v^2,\ p_v^3,\ \cdots,\ p_v^{N_v}}_{\text{CLIP visual tokens}~\bm{p}_v}\ ],
\label{eq:token}
% \vspace{-9pt}
\end{equation}
where $(p_l^1,\ p_l^2,\ \cdots,\ p_l^{N_l})$ are the CLIP language tokens from $\bm{p}_l$, $(cls^1,\ p_v^2,\ p_v^3,\ \cdots,\ p_v^{N_v})$ are the CLIP visual token from $\bm{p}_v$, $[cls]$ represents the classification token generated by CLIP image encoder. $p_r$ represents $[Reg]$ token\cite{deng2021transvg}, which is used to output the region box regression results, and it is randomly initialized and optimized with the whole model. The final one used for regressing the bounding box is a multi-layer perceptron (MLP), which is a three-layer feedforward network, each consisting of a linear layer and a ReLU activation layer.

To prevent catastrophic forgetting and maintain the generalization ability of CLIP, we freeze the parameters of CLIP encoders during training, so that we only need to adapt a few parameters. CLIP-VG does not use any whistle and bells (\textit{e.g.}, ResNet, Cross-attention\cite{jiang2022pseudo}, Query shifts\cite{ye2022shifting}, \etc ~in the visual grounding SOTA models).

\vspace{-2pt}
\subsection{Reliability Measurement}
\vspace{-2pt}
\label{3.3_reliability}

Our approach builds upon the general curriculum learning paradigm\cite{bengio2009curriculum}, where a model goes through multiple rounds of easy-to-hard training by leveraging its own past predictions. In order to facilitate the unsupervised transfer in the grounding task, we utilize a model that has been trained on the original pseudo-labels to apply a pseudo-label quality measurement for selecting the subset of pseudo-labels and then iteratively repeating this process in a self-training cycle.

In uni-modal tasks, the difficulty of the data can be easily measured by predefined rules, such as sentence length, Part Of Speech entropy in NLP, number of objects in CV, \etc \cite{wang2021survey} However, due to the semantic correlation of cross-modal grounding data, the quality of pseudo-labels in visual grounding cannot be evaluated directly. Thus, we define a measurement to evaluate the pseudo-label quality, named \textbf{\textit{Reliability}}, which is calculated as the likelihood of being correctly predicted by the grounding model that is learned with a specific label source.  We believe that, the higher the Reliability, the closer the pseudo-label is to the correct label, rather than noise or unreliable data.

In the case of single-source, in order to acquire the specific Reliability of each pseudo-triplet sample, we define a preliminary grounding model directly learned from all the pseudo-labels, as \textit{\textbf{Reliability Measurer}} $\mathcal{M}$:
% \vspace{-5pt}
\begin{equation}
% \vspace{\eqmargin}
% \vspace{\eqmargin}
\mathcal{M}=\arg \min _{\mathcal{F}_{\theta}} \ell\big(\mathcal{F}_{\theta}(\mathcal{I},\mathcal{E}),\mathcal{B}\big),
\label{eq:measurer_s}
% \vspace{-5pt}
\end{equation}

\noindent and define the \textit{\textbf{Reliability}} $\mathscr{r}$ of a single sample as: 
% \vspace{-2pt}
\begin{equation}
% \small
{\mathscr{r}}=\text{IOU}(\mathcal{M}(\mathscr{i},\mathscr{e}),\mathscr{b}),~~ r\in[0,1.0]
\label{eq:r}
% \vspace{-2pt}
\end{equation}
where $\mathscr{i},\mathscr{e},\mathscr{b}$ represents the \textit{image}, \textit{expression text}, and \textit{bbox} in a pseudo-triplet-paired sample. The IOU is a metric function that can compute the Jaccard overlap between the predicted box and the pseudo box for each sample. Then, we can compute the \textit{\textbf{set of Reliability}} $\mathscr{R}$ for all samples as follows:
\vspace{-2pt}
\begin{equation}
{\mathscr{R}}=\text{IOU}(\mathcal{M}(\mathcal{I},\mathcal{E}),\mathcal{B}).
\label{eq:R_s}
% \vspace{-5pt}
\end{equation}

When considering the multi-source case, we define a \textbf{\textit{group of Reliability Measurers}} $\{\mathcal{M}_{i}\}_{i=1}^n$, where each of them is learned from a specific pseudo-label source:
% \vspace{-3pt}
\begin{equation}
% \vspace{\eqmargin}
% \vspace{\eqmargin}
\mathcal{M}_{i}=\arg \min _{\mathcal{F}_{\theta}} \ell\big(\mathcal{F}_{\theta}(\mathcal{I},\mathcal{E}_i),\mathcal{B}_i\big).
\label{eq:measurer_m}
% \vspace{\eqmargin}
\end{equation}
\noindent Similarly, the \textbf{\textit{set of Reliability}} $\mathscr{R}_{ij}$ is defined as:
% \vspace{-3pt}
\begin{equation}
{\mathscr{R}_{ij}}=\text{IOU}(\mathcal{M}_{i}(\mathcal{I},\mathcal{E}_j),\mathcal{B}_j), \ i \in [1,n], \ j \in [1,n],
\label{eq:R_m}
% \vspace{-5pt}
\end{equation}
\noindent where ${\mathscr{R}_{ij}}$ denotes the set of reliability values for all samples in the $j$-th data source obtained by the $i$-th measurer $\mathcal{M}_{i}$. The ${\mathscr{R}_{ij}}$ denotes \textit{\textbf{Source-specific Reliability}} (SR) when $i=j$ or \textit{\textbf{Cross-source Reliability}} (CR) when $i\neq j$. 

\vspace{+2pt}
{\bf Reliability Histogram.} In order to facilitate the pseudo-label sampling during self-paced curriculum learning, 
we define \textbf{\textit{Reliability Histogram}} (RH) $\mathscr{H}$ or $\mathscr{H}_{ij}$ for each pseudo-label source based on the corresponding set of Reliability $\mathscr{R}$ or $\mathscr{R}_{ij}$ in the single-source or multi-source case. The RH (\eg, \cref{fig:sup_reliability}) has $m$ bins covering the range of Reliability, and each bin represents the number of samples with the reliability value in the corresponding bin interval.
%

%%%%%%%%%%%%%%%%%%%%---------------------------------------------------

\vspace{-2pt}
\subsection{Single-source Self-paced Adapting (SSA)}
% \vspace{-2pt}
\label{3.4SSA}

\begin{algorithm}[t]  % 如果用了 algorithm 包就需要用 [H]
    \label{algo:algo1}
    \SetAlgoLined
    \LinesNumbered
    \begin{small}
    \caption{Single-source Self-paced Adapting(SSA)}
    \KwIn{\textit{pseudo triplet-paired data} $\mathcal{D}$.} 
    \KwOut{\textit{subset} $\mathcal{D}_\chi^*$ \textit{, well-trained optimal model} $\mathcal{F}_{\theta}^*$.} 
    % \BlankLine
    \textbf{\textit{Training Reliability Measurer}} \textit{$\mathcal{M}$}: \\
    \mbox{}\quad $\mathcal{M}$ ← training $\mathcal{F}_\theta^{*}$ in $\mathcal{D}$ by using \cref{eq:measurer_s}\; 
    \textbf{\textit{Sorted Data by Reliability}}: \\
    \mbox{}\quad $\mathcal{H}$←$\mathcal{M}$ measure $\mathcal{D}$ by using \cref{eq:R_s}\;
    \While{\textbf{Curriculum Scheduler}} {
        \textbf{set}\textit{ $\mathscr{h}_0=0.50, \mathrm{\Delta}=0.10$},\ $\mathscr{h}_m=\mathscr{h}_0$, init $\mathcal{D}_\chi=null$\;
        \textbf{\textit{Greedy sample selection strategy}}:\\
        \While{$\mathscr{h}_m$ not optimal} {
            $\mathscr{h}_r=\mathscr{h}_{m}+\Delta$, $\mathscr{h}_l=\mathscr{h}_{m}-\Delta$\;
            training model $\mathcal{F}_\theta^{*}$ for $\mathscr{h}_m$, $\mathscr{h}_r$, $\mathscr{h}_l$ by \cref{eq:model_s}\;
            greedily update $\mathscr{h}_m = \mathscr{h}_l\ or\ \mathscr{h}_r$ by binary search\;
        }
        $\mathscr{h}^{*} = \mathscr{h}_m$, abtain $\mathcal{D}_\chi^*$, model $\mathcal{F}_\theta^{*}$ by \cref{eq:h_s,eq:model_s}\;
    }
    \Return $\mathcal{D}_\chi^*$ and $\mathcal{F}_{\theta}^*$.
    \end{small}
\end{algorithm}

\begin{algorithm}[t] 
    \label{algo:algo2}
    \SetAlgoLined
    \LinesNumbered
    \begin{small}
    \caption{Multi-source Self-paced Adapting(MSA)}
    \KwIn{\textit{multi-source pseudo triplet-paired data} $\mathcal{S}_i,\ i\in\left[1,n\right]$.} 
    \KwOut{\textit{subset} $\mathcal{D}_\chi^*$ \textit{, well-trained optimal model} $\mathcal{F}_{\theta}^*$.} 
    % \BlankLine
    \If{\textbf{\textit{Pseudo-Label Source Selection}}} {
    \textbf{For} $i\ in\ [1, n]$\ \textbf{do}: Compute $average\ entities$ in $\mathcal{S}_i$\; 
% 		using \cref{eq:SourceReliability}
    Reorder $\mathcal{S}_1,\mathcal{S}_2,\dots,\mathcal{S}_n$ according $average\ entities$\;
    }
    \textit{\textbf{Training Reliability Measurer}\ $\mathcal{M}_i$}\ : \\
    \mbox{}\quad \textbf{For} $i\ in\ [1,n]$\ \textbf{do}: $\mathcal{M}_i$ ← training $\mathcal{F}_\theta^{*}$ in $\mathcal{S}_i$ by \cref{eq:measurer_m}\;
    \While{\textbf{Sorted Data by Reliability}} {
        \For{$i\ in\ [1,n]$, $j\ in\ [1,n]$} {
            $\mathcal{H}_{ij}$←$\mathcal{M}_{i}$ measure $\mathcal{S}_j$ by using \cref{eq:R_m}\;
        }
    }
    \While{\textbf{Curriculum Scheduler}} {
        \For{$\mathcal{S}_1 : \mathcal{S}_n$} {
        \textbf{set}\textit{ $\mathscr{h}_0=0.50, \mathrm{\Delta}=0.10$},\ $\mathscr{h}_m=\mathscr{h}_0$, init $\mathcal{D}_\chi=null$\;
        \textit{\textbf{Reliability Measurer Selection}}:\\
        determine best reliability measurer $\mathcal{M}_{i^*}$ by \cref{eq:select_i}\;
        \textbf{\textit{Greedy sample selection strategy}}:\\
        \While{ $\mathscr{h}_m$ not optimal} {
	    $\mathscr{h}_r=\mathscr{h}_{m}+\Delta$, $\mathscr{h}_l=\mathscr{h}_{m}-\Delta$\;
		training model $\mathcal{F}_\theta^{*}$ for $\mathscr{h}_m$, $\mathscr{h}_r$, $\mathscr{h}_l$ by \cref{eq:model_m}\;
		greedily update $\mathscr{h}_m=\mathscr{h}_l\ or\ \mathscr{h}_r$ by binary search\;
        }
        set $\mathscr{h}^* = \mathscr{h}_m$\;
        update $\mathcal{D}_\chi$, and model $\mathcal{F}_\theta^{*}$ by using \cref{eq:D_x,eq:h_m,eq:final-model} 
        }
    }
    \Return $\mathcal{D}_\chi^*$ and $\mathcal{F}_{\theta}^*$.
    \end{small}
\end{algorithm}

To achieve a stable adaption of the CLIP-based network architecture by finding reliable pseudo-labels, we propose the Single-source Self-paced Curriculum Adapting algorithm (SSA) to gradually sample reliable triplet-paired pseudo-labels with a careful curriculum choice based on the reliability measurement. The pipeline and formulation of SSA are shown in \cref{fig:fig4} and Algorithm 1.  % \cref{algo:algo1}

We first train a reliability measurer $\mathcal{M}$ for all single-source pseudo-labels in a self-training manner, and then score the reliability for all samples based on the learned measurer. According to the Reliability results $\mathcal{R}$, a reliability histogram $\mathcal{H}$ (\eg, \cref{fig:sup_reliability}\textcolor{red}{-(a1)}) is constructed to complete the sorting of the pseudo-labels. The follow-up work is to find the pseudo-labels that can optimize the model performance according to the reliability histogram.

To facilitate sampling, we define a reliability threshold $\mathscr{h}$, and use it to sample a subset from the pseudo-label source. Specifically, we define $percent\left(\mathcal{H},\mathscr{h}\right)$ as the extracted subset from the current pseudo-label source according to reliability histogram $\mathcal{H}$, where each sample has the reliability value belongs to the interval $[\mathscr{h}, 1.0]$. The number of samples in the subset can be computed mathematically as:
% \vspace{-2pt}
\begin{equation}
% \vspace{\eqmargin}
\left|percent\left(\mathcal{H},\mathscr{h}\right)\right|={\textstyle \sum_{r=\mathscr{h}_0}^{1.0}}\mathcal{H}\left(r\right).
\label{eq:percent_s}
\end{equation}

\noindent Particularly, when $\mathscr{h}=0$, all data is selected. Then, the goal is to find the optimal Reliability threshold $\mathscr{h^*}$ with the best performance on the validation set:
% \vspace{-5pt}
\begin{equation}
% \vspace{-5pt}
h^*=\arg \min _{h,\mathcal{F}_{\theta}} \ell\left(\mathcal{F}_{\theta}(percent\left(\mathcal{H},\mathscr{h}\right))\right).
\label{eq:h_s}
% \vspace{-2pt}
\end{equation}

\vspace{2pt}
\textbf{Greedy Sample Selection.} The cost is unbearable if the threshold $\mathscr{h}$ is traversed over the [0, 1.0] interval. Therefore, we propose a greedy sample selection strategy based on the modified binary search. Specifically, we define $\mathscr{h}_r$, $\mathscr{h}_m$ and $\mathscr{h}_l$ as three temporary thresholds. It is worth noting that the experimental results show that the model performance usually tends to saturate around the reliability threshold $\mathscr{h}=0.5$. Thus, we initialize the $\mathscr{h}_m$ as $0.5$, and fix $\mathscr{h}_r=\mathscr{h}_{m}+\Delta$ while $\mathscr{h}_l=\mathscr{h}_{m}-\Delta$. Then, greedily solving Eq.~\eqref{eq:h_s} by trying different values of $\mathscr{h}_m$. We keep updating $\mathscr{h}_m=\mathscr{h}_l\ or\ \mathscr{h}_r$ until $\mathscr{h}_m$ achieves better performance than both $\mathscr{h}_l$ and $\mathscr{h}_r$. Based on this strategy, we can quickly find the appropriate reliability threshold with sub-optimal performance, thus reducing the model training cost and ensuring a balance between reliable and unreliable samples.

\subsection{Multi-source Self-paced Adapting (MSA)}
\label{3.5MSA}

Since the proposed self-paced adapting algorithm does not depend on the specific form or quality of the pseudo-labels, it can be flexibly extended to access multiple sources of pseudo-labels. Using multiple sources of pseudo-labels will increase the diversity of language taxonomy and match patterns, as well as the richness of contextual semantics, thus improving the generalization ability of the visual grounding model. In real scenarios, obtaining multiple sources of pseudo-language labels from various vision and language contexts is not difficult (\eg, large-scale corpus, visual question answering, image captioning, scene graph generation, visual language navigation, \etc). We will introduce the details of how to obtain multiple sources of pseudo-language labels in \cref{4.1detail}.

The impact of unreliable data will be more severe with the inclusion of multi-source pseudo-labels. Moreover, resolving this issue is not easy due to the distribution discrepancy in language taxonomy among different label sources. Therefore, we propose Multi-source self-paced Adapting (MSA) based on SSA, as depicted in \cref{fig:fig4} and Algorithm 2.

\vspace{+2pt}
\textbf{Pseudo-Label Source Selection.} 
Before the execution of MSA, we need to decide which label source to be used for adapt training. We propose to compute \textit{\textbf{the average number of entities per expression}} in each label source as the difficulty criterion at source level, which can be used to sort the label sources from simple to complex. We assume that the selected data source is $\mathcal{S}_{j^*}$ in the current MSA step. Then, we can gradually consider one label source from simple to complex for learning the grounding model in each step of the MSA. 

\vspace{+2pt}
\textbf{Reliability Measurer Selection.}
Reliability measures learned from different pseudo sources exhibit divergent discriminative abilities for a given source. As introduced in \cref{3.3_reliability}, we can obtain multiple Reliability (\textit{i.e.}, $\{\mathscr{R}_{ij^*}\}_{i=1}^n$) for the data source $\mathcal{S}_{j^*}$ obtained by different Reliability Measurers. Therefore, we need to select an optimal Reliability Measurer for sampling pseudo-labels from the data source used in the current MSA step.
 
We firstly set a Reliability threshold $\mathscr{h}_0$ (\eg, generally $\mathscr{h}_0=0.5$), and use it to select a subset of the pseudo samples from the current data source. Specifically, we define $percent\left(\mathcal{H}_{ij^*},\mathscr{h}_0\right)$ as the extracted subset from the $j^*$-th data source according to $\mathcal{H}_{ij^*}$. The calculation of the samples' number is similar as \cref{eq:percent_s}, that is:
% \vspace{-3pt}
\begin{equation}
% \vspace{\eqmargin}
\left|percent\left(\mathcal{H}_{ij^*},\mathscr{h}_0\right)\right|=
 {\textstyle \sum_{r=\mathscr{h}_0}^{1.0}}\mathcal{H}_{ij^*}\left(r\right).
\label{eq:percent_m}
 % \vspace{-0pt}
\end{equation}

Next, we choose the optimal Reliability Measurer $\mathcal{M}_{i^*}$ with the best performance on the validation set by conducting model training and validation after adding the selected subset to $\mathcal{D}_\chi$ (\cref{eq:percent_m}):
% \vspace{-2pt}
\begin{equation}
\vspace{\eqmargin}
i^*=\arg \min _{i,\mathcal{F}_{\theta}} \ell\left(\mathcal{F}_{\theta}(\mathcal{D}_\chi\cup percent\left(\mathcal{H}_{ij^*},\mathscr{h}_0\right))\right),
\label{eq:select_i}
% \vspace{-2pt}
\end{equation}
where $\mathcal{D}_\chi$ is the whole subset of selected pseudo samples before the current MSA step, which is initiated with $null$. 

\vspace{+2pt}
\textbf{Greedy Sample Selection.}
After determining the optimal Reliability Measurer $\mathcal{M}_{i^*}$, we further select pseudo samples from the current data source $\mathcal{S}_{j^*}$ according the the corresponding Reliability Histogram $\mathcal{H}_{i^*j^*}$. Specifically, we find the optimal Reliability threshold $\mathscr{h^*}$ with the best performance on the validation set:
\vspace{-2pt}
\begin{equation}
\vspace{\eqmargin}
h^*=\arg \min _{h,\mathcal{F}_{\theta}} \ell\left(\mathcal{F}_{\theta}(\mathcal{D}_\chi\cup percent\left(\mathcal{H}_{i^*j^*},\mathscr{h}\right))\right).
\label{eq:h_m}
% \vspace{-2pt}
\end{equation}
\noindent This step also adopts the greedy sample selection, which is the same as the SSA in \cref{3.4SSA}. Then, we select the pseudo samples with reliability values in the interval $[\mathscr{h^*}, 1.0]$ from histogram $\mathcal{H}_{i^*j^*}$. Finally, we add the selected pseudo samples to the whole sample set $\mathcal{D}_\chi$ as follows: 
\vspace{-2pt}
\begin{equation}
% \vspace{\eqmargin}
\mathcal{D}_\chi = \mathcal{D}_\chi\cup percent\left(\mathcal{H}_{i^*j^*},\mathscr{h}^*\right).
\label{eq:D_x}
% \vspace{-2pt}
\end{equation}

At the end of the self-paced learning, we will obtain a final subset of pseudo-labels $\mathcal{D}_\chi^*$, which can be utilized to learn the ultimate grounding model:
\vspace{-3pt}
\begin{equation}
% \vspace{-5pt}
\mathcal{F}_{\theta}^{*}=\arg \min _{\mathcal{F}_{\theta}} \ell\left(\mathcal{F}_{\theta}(\mathcal{D}_\chi^*)\right).
\label{eq:final-model}
% \vspace{-3pt}
\end{equation}

%%%%%%%%%%%%%%%%%%%%%%%%%%%%%%%%%%%%%%%%%%%%%%%%%%%%%%%%%%%%%%%%%%%%%%%%%
%------------------------------------------------------------------------
%------------------------------------------------------------------------
%------------------------------------------------------------------------
%------------------------------------------------------------------------
\vspace{\eqmargin}
\section{Experiments}
% \vspace{-3pt}

\subsection{Implementation Details}
\label{4.1detail}
% \vspace{-3pt}

\begin{figure}[t]
\centering
   \includegraphics[width=0.80\linewidth]{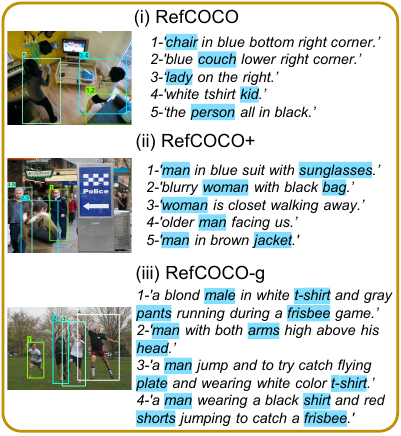}
   \vspace{-8pt}
   \caption{The samples of the validation split in the RefCOCO/+/g dataset. The figure illustrates the characteristics of ground-truth query labels and grounding difficulty among the three datasets, with language entities highlighted in \textcolor{Cyan}{cyan}.} % \textcolor{Turquoise}{Turquoise}
   \label{fig:samples}
\vspace{-8pt}
\end{figure}

\noindent\textbf{Datasets and Settings.} Following previous fully supervised and unsupervised visual grounding work, we evaluate our approach on five mainstream datasets: RefCOCO \cite{yu2016modeling}, RefCOCO+ \cite{yu2016modeling}, RefCOCOg \cite{mao2016generation}, ReferItGame \cite{kazemzadeh2014referitgame}, and Flickr30K Entities \cite{plummer2015flickr30k}. \cref{fig:samples} displays the validation samples in the RefCOCO/+/g dataset. The ground-truth query labels' language characteristics and grounding difficulties differ across the three datasets. The language complexity of RefCOCO/+/g increases with the number of language entities. In our experiments, we adopt exactly the same train/val/test image splits as in TransVG \cite{deng2021transvg} and Pseudo-Q \cite{jiang2022pseudo}. The number of training images in the five datasets is 16,994, 16,992, 24,698, 8,994, and 29,779, respectively. It should be noted that in the unsupervised setting, we do not use any manually labeled data or bounding boxes as supervised information during training, which is only used for testing purposes.

\vspace{+2pt}
\noindent\textbf{Sources of Different Pseudo-Language Labels.} In the case of single-source (\cref{3.4SSA}), we utilize template pseudo-labels that are generated by the generation module in Pseudo-Q\cite{jiang2022pseudo}. These labels are synthesized from spatial relationship prior knowledge and object labels provided by the detectors, which include category and attribute information. For instance, one of the templates and examples is like \textit{$\{$Relation-Object-Attribute$\}$, “right man standing”}. However, template pseudo-labels lack grammatical and logical structures, while language taxonomy is limited by detector-recognized categories. As illustrated in \cref{fig:samples}, this poses a challenge for the model to acquire more sophisticated language logic and semantic comprehension abilities. Therefore, exploiting multi-source pseudo-language labels becomes imperative for unsupervised grounding tasks.

In the case of multi-source (\cref{3.5MSA}), in addition to template pseudo-labels (abbreviate as \textit{tmp.}), we utilized RelTR\cite{cong2022reltr} based on the scene graph generation (SGG) to generate scene graph relation as the pseudo-relation label (abbreviate as \textit{rel.}), and utilized M2\cite{cornia2020meshed} / CLIPCap\cite{mokady2021clipcap} based on the image captioning (IC) to generate caption as the pseudo-caption label (abbreviate as \textit{cap.}) (as shown in \cref{fig:fig4}\textcolor{red}{-(a)}). As for the pseudo bbox, the paired bbox of the subject in SGG is used for the pseudo-relation label, while for the pseudo-caption label, we obtain its pseudo bbox by utilizing an NLP parser (\eg, spaCy) to extract the subject and then pairing it with a bbox provided by the same detectors. However, these pseudo-labels also contain a significant amount of unreliable samples and noise. It is worth noting that these pseudo-label sources are only used to validate the effectiveness of our algorithm, and our method is not restricted to these pseudo-language labels.

%%%%%%%%%%%%%%%%%%%%%%%%%%%%%%%%%%%%%%%%%%%%%%%%%%%%%%%%%%%%%%%%%%%%%
\begin{table*}[t]\footnotesize
\vspace{\figmargin}
\caption{Comparison with state-of-the-art methods on RefCOCO~\cite{yu2016modeling}, RefCOCO+~\cite{yu2016modeling} and RefCOCOg~\cite{mao2016generation} datasets in terms of $top$-1 accuracy (\%). \textit{``Un.''} represents unsupervised. \textit{``Sup."} refers to supervision level: \textit{Single-source} and \textit{Multi-source} are the unsupervised cases (without annotation), \textit{Weakly} (only annotated queries), and \textit{Fully} (annotated bbox-query pairs). The best two results with supervision levels of (Single-source unsupervised + Weakly) and Fully are \textbf{bold-faced} and \underline{underlined}, respectively. The ``$\dagger$" in the table indicates that the results of Pseudo-Q in the multi-source scenario are obtained by directly training the Pseudo-Q model on data with mixed pseudo-labels from multiple sources. The results also show the performance gains of the SSA and MSA algorithms. \textit{w/o} represents ‘without’, \textit{w.} represents ‘with’. Our results are highlighted in \colorbox{blue!10}{blue} shading, while the main comparison SOTA models are highlighted in \colorbox{gray!16}{gray} shading.}
\vspace{-12pt}
\begin{center}
\resizebox{1.90\columnwidth}{!}{%
\begin{tabular}{p{2.3cm} | p{1.2cm}<{\centering} | p{0.7cm}<{\centering} | p{0.8cm}<{\centering} p{0.8cm}<{\centering} p{0.8cm}<{\centering} | p{0.8cm}<{\centering} p{0.8cm}<{\centering} p{0.8cm}<{\centering} | p{0.8cm}<{\centering} p{0.8cm}<{\centering} p{0.8cm}<{\centering}}
    \toprule
    % \midrule
\multirow{2}{*}{Method} & \multirow{2}{*}{Venue} & \multirow{2}{*}{Sup.} & \multicolumn{3}{c|}{RefCOCO} & \multicolumn{3}{c|}{RefCOCO+} & \multicolumn{3}{c}{RefCOCOg} \\
 & & & val & testA & testB & val & testA & testB & val-g & val-u & test-u \\
    \midrule
    \midrule
CPT \cite{yao2021cpt} & \textit{arXiv'21} & \multirow{4}{*}{\textit{\makecell{Un.\\ Single- \\ source }}} & 32.20 & 36.10 & 30.30 & 31.90 & 35.20 & 28.80 & - & 36.70 & 36.50 \\
\cellcolor{gray!12} Pseudo-Q \cite{jiang2022pseudo} & \cellcolor{gray!12} \textit{CVPR'22} & & \cellcolor{gray!12} \underline{56.02} & \cellcolor{gray!12}\underline{58.25} & \cellcolor{gray!12}\underline{54.13} & \cellcolor{gray!12}38.88 & \cellcolor{gray!12}\underline{45.06} & \cellcolor{gray!12}32.13 & \cellcolor{gray!12}\underline{49.82} & \cellcolor{gray!12}\underline{46.25} & \cellcolor{gray!12}\underline{47.44} \\ 
CLIP-VG \textit{w/o} SSA & \textit{--} & & 57.92 & 61.92 & 54.82 & 47.92 & 52.07 & 35.21 & 52.31 & 51.45 & 51.47 \\  % \rowcolor{gray!20}
\cellcolor{blue!9} \textbf{CLIP-VG} (Ours) & \cellcolor{blue!9} \textit{TMM'23} &  & \cellcolor{blue!9} \textbf{62.38} & \cellcolor{blue!9} \textbf{65.03} & \cellcolor{blue!9} \textbf{56.64} & \cellcolor{blue!9} \textbf{48.87} & \cellcolor{blue!9} \textbf{55.73} & \cellcolor{blue!9} \textbf{39.41} & \cellcolor{blue!9} \textbf{54.16} & \cellcolor{blue!9} \textbf{54.11} & \cellcolor{blue!9} \textbf{54.81} \\
    % \midrule
    \midrule[0.005pt]  % 控制粗细
\cellcolor{gray!12} Pseudo-Q$^\dagger$  & \cellcolor{gray!12}\textit{--} & \multirow{3}{*}{\textit{\makecell{Un.\\ Multi- \\ source }}} & \cellcolor{gray!12}50.23 & \cellcolor{gray!12}54.38 & \cellcolor{gray!12}48.25 & \cellcolor{gray!12}37.25 & \cellcolor{gray!12}42.44 & \cellcolor{gray!12}31.87 & \cellcolor{gray!12}47.32 & \cellcolor{gray!12}45.86 & \cellcolor{gray!12}46.12 \\ 
CLIP-VG \textit{w/o} MSA & \textit{--} & & 60.18 & 65.04 & 57.03 & 48.23 & 54.21 & 38.39 & 55.26 & 54.54 & 54.44 \\  % \rowcolor{gray!20}
\cellcolor{blue!9} \textbf{CLIP-VG} (Ours) & \cellcolor{blue!9} \textit{TMM'23} &  & \cellcolor{blue!9} \textbf{64.89} & \cellcolor{blue!9} \textbf{69.03} & \cellcolor{blue!9} \textbf{59.12} & \cellcolor{blue!9} \textbf{50.85} & \cellcolor{blue!9} \textbf{57.31} & \cellcolor{blue!9} \textbf{41.27} & \cellcolor{blue!9} \textbf{58.06} & \cellcolor{blue!9} \textbf{56.54} & \cellcolor{blue!9} \textbf{57.51} \\
    \midrule
% VC \cite{zhang2018grounding} & \textit{CVPR'18} & \multirow{4}{*}{\textit{Weak}} & - & 33.29 & 30.13 & - & 34.60 & 31.58 & 33.79 & - & - \\
ARN \cite{liu2019adaptive} & \textit{ICCV'19} & \multirow{3}{*}{\textit{Weakly}} & 34.26 & 36.43 & 33.07 & 34.53 & 36.01 & 33.75 & 33.75 & - & - \\
KPRN \cite{liu2019knowledge} & \scriptsize{\textit{ACMMM'19}} & & 35.04 & 34.74 & 36.98 & 35.96 & 35.24 & 36.96 & 33.56 & - & - \\
DTWREG \cite{sun2021discriminative} & \textit{TPAMI'21} & & 39.21 & 41.14 & 37.72 & \underline{39.18} & 40.10 & \underline{38.08} & 43.24 & - & - \\
    \midrule
% MAttNet \cite{yu2018mattnet} & \textit{CVPR'18} & \multirow{9}{*}{\textit{Full}} & 76.65 & 81.14 & 69.99 & 65.33 & 71.62 & 56.02 & - & 66.58 & 67.27 \\
% NMTree \cite{liu2019learning} & \textit{ICCV'19} & & 76.41 & 81.21 & 70.09 & 66.46 & 72.02 & 57.52 & 64.62 & 65.87 & 66.44 \\
% FAOA \cite{yang2019fast} & \textit{ICCV'19} & & 72.54 & 74.35 & 68.50 & 56.81 & 60.23 & 49.60 & 56.12 & 61.33 & 60.36 \\
% ReSC \cite{yang2020improving} & \textit{ECCV'20} & & 77.63 & 80.45 & 72.30 & 63.59 & 68.36 & 56.81 & 63.12 & 67.30 & 67.20 \\
TransVG \cite{deng2021transvg} & \textit{ICCV'21} &  \multirow{5}{*}{\textit{Fullly}}  & 80.83 & 83.38 & 76.94 & 68.00 & 72.46 & 59.24 & 68.03 & 68.71 & 67.98 \\
Refformer \cite{li2021referring} & \textit{NIPS'21}  & & 82.23 & 85.59 & 76.57 & 71.58 & 75.96 & \underline{62.16} & - & 69.41 & 69.40 \\
VGTR    \cite{du2022visual}    & \textit{ICME'22} & & 79.30 & 82.16 & 74.38 & 64.40 & 70.85 & 55.84 & 64.05 & 66.83 & 67.28 \\
\cellcolor{gray!12} QRNet \cite{ye2022shifting} & \cellcolor{gray!12}\textit{CVPR'22} & & \cellcolor{gray!12}\underline{84.01} & \cellcolor{gray!12}\underline{85.85} & \cellcolor{gray!12}\textbf{82.34} & \cellcolor{gray!12}\textbf{72.94} & \cellcolor{gray!12}\underline{76.17} & \cellcolor{gray!12}\textbf{63.81} & \cellcolor{gray!12}\underline{71.89} & \cellcolor{gray!12}\underline{73.03} & \cellcolor{gray!12}\underline{72.52} \\ % \rowcolor{gray!12}
\cellcolor{blue!9} \textbf{CLIP-VG} (Ours) & \cellcolor{blue!9} \textit{TMM'23} & & \cellcolor{blue!9} \textbf{84.29} & \cellcolor{blue!9} \textbf{87.76} & \cellcolor{blue!9} \underline{78.43} & \cellcolor{blue!9} \underline{69.55} & \cellcolor{blue!9} \textbf{77.33} & \cellcolor{blue!9} 57.62 & \cellcolor{blue!9} \textbf{72.64} & \cellcolor{blue!9} \textbf{73.18} & \cellcolor{blue!9} \textbf{72.54} \\
    \bottomrule
\end{tabular}%
}
\end{center}
\label{tab:SOTA1}
\vspace{-15pt}	
\end{table*}

%%%%%%%%%%%%%%%%%%%%%%%%%%%%%%%%%%%%%%%%%%%%%%%%%%%%%%%%%%%%%%%%
\begin{table}[t]\footnotesize
\caption{Comparison with state-of-the-art methods on ReferItGame~\cite{kazemzadeh2014referitgame} and Flickr30K Entities~\cite{plummer2015flickr30k} in terms of $top$-1 accuracy (\%) in test split. Annotations are the same as \cref{tab:SOTA1}.}
\vspace{-10pt}
\begin{center}
\resizebox{0.96\columnwidth}{!}{%
\begin{tabular}{p{2.3cm} | p{1.0cm}<{\centering} | p{1.0cm}<{\centering} | p{1cm}<{\centering} | p{1cm}<{\centering}}
    \toprule
    % \midrule
Method & Venue & \textit{Sup.} & ReferIt & Flickr30K \\
    \midrule
    \midrule
% Yeh \textit{et al}.~\cite{yeh2018unsupervised} & \textit{CVPR'18} & \multirow{2}{*}{\textit{Un.}} & 36.93 & 20.91 \\
Wang \textit{et al}.~\cite{wang2019phrase} & \textit{ICCV'19} & \multirow{5}{*}{\textit{\makecell{Un.\\ Single- \\ source}}} & 26.48 & 50.49 \\
BiCM \cite{shi2022unpaired} & \textit{arXiv'22} &  & 42.96 & \underline{61.46}  \\
\cellcolor{gray!12} Pseudo-Q \cite{jiang2022pseudo} & \cellcolor{gray!12} \textit{CVPR'22} &   & \cellcolor{gray!12} \underline{43.32} & \cellcolor{gray!12} 60.41  \\ 
CLIP-VG \textit{w/o} SSA & \textit{--} &   & 46.16 &  61.66 \\ % \rowcolor{gray!12}
\cellcolor{blue!9} \textbf{CLIP-VG} (Ours) & \cellcolor{blue!9} \textit{TMM'23} &  & \cellcolor{blue!9} \textbf{50.63} & \cellcolor{blue!9} \textbf{64.51}  \\
    % \midrule
    \midrule[0.005pt]  % 控制粗细
    % \hdashline[3pt/5pt]
    % \hdashline
\cellcolor{gray!12} Pseudo-Q$^\dagger$ & \cellcolor{gray!12} \textit{--} & \multirow{3}{*}{\textit{\makecell{Un.\\ Multi- \\ source}}} & \cellcolor{gray!12} 42.31 & \cellcolor{gray!12} 56.77  \\
CLIP-VG \textit{w/o} MSA & \textit{--} &  & 49.30 & 63.33  \\ % \rowcolor{gray!12}
\cellcolor{blue!8} \textbf{CLIP-VG} (Ours) & \cellcolor{blue!8} \textit{TMM'23} &  & \cellcolor{blue!8} \textbf{53.08} & \cellcolor{blue!9} \textbf{66.62}  \\
    \midrule
% Chen \textit{et al}.~\cite{chen2018knowledge} & \textit{CVPR'18} & \multirow{6}{*}{\textit{Weak}} & 33.67 & 46.61 \\
% Zhao \textit{et al}.~\cite{zhao2018weakly} & \textit{CVPR'18} & & 33.10 & 13.61 \\
% Liu \textit{et al}.~\cite{liu2019adaptive} & \textit{ICCV'19} & & 26.19 & - \\
Gupta \textit{et al}.~\cite{gupta2020contrastive} & \textit{ECCV'20} & \multirow{3}{*}{\textit{Weakly}} & - & 51.67  \\
Liu \textit{et al}.~\cite{liu2021relation} & \textit{CVPR'21} & & 37.68 & 59.27  \\
Wang \textit{et al}.~\cite{wang2021improving} & \textit{CVPR'21} & & 38.39 & 53.10  \\
    \midrule
% Kovvuri \textit{et al}.~\cite{kovvuri2018pirc} & \textit{ACCV'18} & \multirow{7}{*}{\textit{Full}} & 59.13 & 72.83 \\
% Yu \textit{et al}.~\cite{yu2018rethinking} & \textit{IJCAI'18} & & 63.00 & 73.30 \\
% Yang \textit{et al}.~\cite{yang2019fast} & \textit{ICCV'19} & & 60.67 & 68.71 \\
% Yang \textit{et al}.~\cite{yang2020improving} & \textit{ECCV'20} & & 64.60 & 69.28 \\
TransVG~\cite{deng2021transvg} & \textit{ICCV'21} & \multirow{5}{*}{\textit{Fullly}} & 70.73 & 79.10 \\
Refformer~\cite{li2021referring} & \textit{NIPS'21} & & 70.81 & 78.13 \\
VGTR~\cite{du2022visual} & \textit{ICME'22} & & - & 74.17 \\
\cellcolor{gray!12} QRNet \cite{ye2022shifting} & \cellcolor{gray!12}\textit{CVPR'22} & & \cellcolor{gray!12}\textbf{74.61} & \cellcolor{gray!12}\underline{81.95}  \\ % \rowcolor{gray!12}
\cellcolor{blue!9} \textbf{CLIP-VG} (Ours) & \cellcolor{blue!9} \textit{TMM'23} & & \cellcolor{blue!9} \underline{70.89} & \cellcolor{blue!9} \textbf{81.99}  \\
    \bottomrule
\end{tabular}%
}
\end{center}
\label{tab:SOTA2}
\vspace{-17pt}	
\end{table}

%%%%%%%%%%%%%%%%%%%%%%%%%%%%%%%%%%%%%%%%%%%%%%%%%%%%%%%%%%%%%%%%
\begin{table*}[t]\footnotesize
\caption{Training/inference cost comparison. The results are obtained on RefCOCO dataset ($FPS: images / (GPU \cdot second)$).}
\vspace{-12pt}
\begin{center}
\resizebox{1.8\columnwidth}{!}{%
\begin{tabular}{p{2.0cm}<{\centering} | p{1.8cm}<{\centering} | p{1.8cm}<{\centering} | p{1.5cm}<{\centering} | p{0.7cm}<{\centering} | p{0.7cm}<{\centering} | p{0.8cm}<{\centering} | p{0.5cm}<{\centering}| p{0.8cm}<{\centering} | p{0.8cm}<{\centering} | p{0.7cm}<{\centering}}
    \toprule
    % \midrule
\multirow{2}{*}{Model} & Vision Backbone & Language Backbone & Cross-modal Backbone & All params. & Update params. & Training FPS & \multirow{2}{*}{epoch} &  GPU hours(h) & Test FPS & testA Time(s)\\
    \midrule
    \midrule
Pseudo-Q\cite{jiang2022pseudo}  & ResNet-50  & BERT-base & DETR-R50    & \textbf{156M}  & 156M  & 36.35   &  20            & 14.7   & 78.57   & 72 s    \\ 
TransVG \cite{deng2021transvg}  & ResNet-101 & BERT-base & DETR-R101   & 170M           & 168M  & 22.85   &  90            & 105.0  & 59.55   & 95 s    \\
MDETR \cite{kamath2021mdetr}    & ResNet-101 & BERT-base & DETR-R101   & 185M           & 185M  & 4.71    &  \textbf{5}    & 28.33  & 19.98   & 283 s   \\  \rowcolor{gray!12}
QRNet \cite{ye2022shifting}     & Swin-S     & BERT-base & None        & 273M           & 273M  & 9.41    &  160           & 453.3  & 50.96   & 111 s   \\  \rowcolor{blue!9}
\textbf{CLIP-VG} (Ours)         & CLIP-ViT-B/16 & CLIP-ViT-B/16 & None & 171M  & \textcolor{black}{\textbf{21M}} & \textcolor{black}{\textbf{252.57}} & 90 & \textcolor{black}{\textbf{10.5}} & \textcolor{black}{\textbf{377.85}} & \textcolor{black}{\textbf{15 s}}\\
    \bottomrule
\end{tabular}%
}
\end{center}
\label{tab:cost}
\vspace{-15pt}	
\end{table*}

\vspace{+2pt}

\noindent\textbf{Network Architecture.} We primarily utilize CLIP ViT-B/16 as the backbone, where the image and text encoder is a 12-layer Transformer. The image encoder of CLIP comprises 12 heads with a hidden dimension of 768, and its output is aligned to 512. The text encoder of CLIP has 8 heads and a hidden dimension of 512. The length of the image encoder's token embedding is 197, while the text encoder's token has a length of 77. The cross-modality Transformer only consists of 6 layers, 8 heads, and a hidden dimension of 512. To achieve multi-level representation perception, we extract intermediate features from layers [1,4,8,12] in the image encoder of CLIP.

\vspace{+2pt}
\noindent\textbf{Inputs.} Previous work set the image size to 640×640 and the maximum expression length to 40. Since our model is based on CLIP, we set the image size to 224×224 and the maximum expression length to 77. Specifically, the long side of the image is resized to 224, while the short side is padded to 224, and the language token is filled with empty tokens when the sentence is insufficient for alignment. 

\vspace{+2pt}
\noindent\textbf{Training Details.} Our framework and experiments are all based on PyTorch by using 8 Nvidia RTX3090 GPUs. Our model is optimized end-to-end with AdamW optimizer. The initial learning rate of the cross-modal grounding module is $2.5\times10^{-4}$. All datasets use a cosine learning rate schedule. Training 90 epochs for all models. The batch size is set as 64. To maintain a fair comparison, other unspecified settings are consistent with Pseudo-Q\cite{jiang2022pseudo} and TransVG\cite{deng2021transvg}.

\vspace{-5pt}
\subsection{Comparison with State-of-the-Art Methods}
\label{4.2results}
\vspace{-2pt}

In this section, we validate our approach on five mainstream benchmarks, RefCOCO/+/g \cite{yu2016modeling, mao2016generation}, ReferitGame \cite{kazemzadeh2014referitgame} and Flickr30K Entities \cite{plummer2015flickr30k}. We apply our approach to single-source pseudo-template labels and multi-source pseudo-language labels to verify the effectiveness of our method in unsupervised settings. Additionally, we compare the current mainstream SOTA models in a fully supervised setting by using manual high-quality triplet-paired annotations to confirm the superiority of our model in terms of both speed and energy efficiency.

%%%%%%%%%%%%%%%%%%%%%%%%%%%%%%

\begin{table}[t]\footnotesize
\vspace{-5pt}
\caption{Ablation study on SSA and MSA algorithms adapting process. \textit{tmp.} represents pseudo-template label, \textit{rel.} represents the pseudo-relation label, \textit{cap.} represents the pseudo-caption label. \textit{Imp.} represents an improvement. M. represents the CLIP-VG model. The result obtained by the CLIP-VG model w/o extracting multi-level features. \textit{ml} represents the CLIP-VG model with multi-level feature perception. }
\vspace{-12pt}
\begin{center}
\resizebox{1.0\columnwidth}{!}{%
\begin{tabular}{p{3.4cm} |  p{0.5cm}<{\centering} p{0.9cm}<{\centering} |   p{0.5cm}<{\centering} p{0.8cm}<{\centering} |  p{0.5cm}<{\centering} p{0.8cm}<{\centering}}
    \toprule
    % \midrule
\multirow{2}{*}{Method} &   \multicolumn{2}{c|}{RefCOCO} & \multicolumn{2}{c|}{RefCOCO+} & \multicolumn{2}{c}{Referit} \\
 & testA & \cellcolor{blue!5}Imp. &  testA  & \cellcolor{blue!5}Imp. & test & \cellcolor{blue!5}Imp.\\
    \midrule
    \midrule
% Pseudo-Q\cite{jiang2022pseudo} (baseline)                            & 58.25 & \textcolor{blue}{-     } & 45.06 & \textcolor{blue}{-     } & 43.32 & \textcolor{blue}{-     }  \\
M. + \textit{tmp.}                                                   & 61.82 & \cellcolor{blue!5}\textcolor{black}{--}     & 49.06 & \cellcolor{blue!5}\textcolor{black}{--}      & 43.16 & \cellcolor{blue!5}\textcolor{black}{--}  \\
M. + \textit{rel.}                                                   & 38.74 & \cellcolor{blue!5}\textcolor{black}{-23.08} & 36.94 & \cellcolor{blue!5}\textcolor{black}{-12.12}  & 25.25 & \cellcolor{blue!5}\textcolor{black}{-17.91}  \\
M. + \textit{cap.}                                                   & 42.20 & \cellcolor{blue!5}\textcolor{black}{-19.62} & 40.03 & \cellcolor{blue!5}\textcolor{black}{-9.03 }  & 24.28 & \cellcolor{blue!5}\textcolor{black}{-18.88}  \\
M. + \textit{tmp.} + \textit{rel.}                                   & 62.26 & \cellcolor{blue!5}\textcolor{black}{-0.26 } & 49.09 & \cellcolor{blue!5}\textcolor{black}{↑0.03 }  & 45.18 & \cellcolor{blue!5}\textcolor{black}{↑2.02 }  \\
M. + \textit{tmp.} + \textit{rel.} + \textit{cap.}                   & 62.51 & \cellcolor{blue!5}\textcolor{black}{↑0.69 } & 46.84 & \cellcolor{blue!5}\textcolor{black}{-2.22 }  & 45.68 & \cellcolor{blue!5}\textcolor{black}{↑2.52 }  \\
M. + \textit{tmp.} + \textit{SSA}                                    & 65.20 & \cellcolor{blue!5}\textcolor{black}{↑3.38 } & 52.67 & \cellcolor{blue!5}\textcolor{black}{↑3.61 }  & 49.89 & \cellcolor{blue!5}\textcolor{black}{↑6.73 }  \\
M. + \textit{tmp.} + \textit{rel.} + \textit{MSA}                    & 67.29 & \cellcolor{blue!5}\textcolor{black}{↑5.47 } & 55.32 & \cellcolor{blue!5}\textcolor{black}{↑6.26 }  & 48.91 & \cellcolor{blue!5}\textcolor{black}{↑5.75 }  \\
M. + \textit{tmp.+rel.+cap.}+ \textit{MSA}                           & 68.35 & \cellcolor{blue!5}\textcolor{black}{↑6.53 } & 56.30 & \cellcolor{blue!5}\textcolor{black}{↑7.24 }  & 51.32 & \cellcolor{blue!5}\textcolor{black}{↑8.16 }  \\  \rowcolor{blue!5}
\textit{ml} M.\textit{+tmp.+rel.+cap.+MSA}                 &  \textbf{69.03} & \textbf{\textcolor{black}{↑7.21 }} & \textbf{57.41} & \textbf{\textcolor{black}{↑8.35 }} & \textbf{53.08} & \textbf{\textcolor{black}{↑9.92 }}  \\
    \bottomrule
\end{tabular}%
}
\end{center}
\label{tab:msa_execution}
\vspace{-12pt}	
\end{table}

\begin{table}[t]\footnotesize
% \vspace{-5pt}
\caption{Ablation study of Source-specific Reliability (SR) and Cross-source Reliability (CR).} 
\vspace{-20pt}
\begin{center}
\resizebox{1.0\columnwidth}{!}{%
\begin{tabular}{p{0.6cm} | p{2.8cm}  |  p{0.5cm}<{\centering} p{0.9cm}<{\centering}  |  p{0.5cm}<{\centering} p{0.9cm}<{\centering}  |  p{0.5cm}<{\centering} p{0.9cm}<{\centering}}
    \toprule
    % \midrule
\multirow{2}{*}{Source} & \multirow{2}{*}{Method} &   \multicolumn{2}{c|}{RefCOCO} & \multicolumn{2}{c|}{RefCOCO+} & \multicolumn{2}{c}{ReferIt} \\
& & testA & \cellcolor{blue!5}Imp. & testA & \cellcolor{blue!5}Imp. &  test & \cellcolor{blue!5}Imp.\\
    \midrule    
    \midrule  
\multirow{3}{*}{\textit{\makecell{multi- \\ source}}}
% & Pseudo-Q (baseline)                            &  54.38 & \textcolor{blue}{---   } &  32.44 &  \textcolor{blue}{---   } &  43.31 &  \textcolor{blue}{---}     \\
& M. + \textit{w/o SR, w/o CR.}                  &  65.04 & \cellcolor{blue!5}\textcolor{black}{--}       &  54.21 &  \cellcolor{blue!5}\textcolor{black}{--}     &  49.30 &  \cellcolor{blue!5}\textcolor{black}{--}  \\
& M + \textit{w. SR, w/o CR.}                    &  67.42 & \cellcolor{blue!5}\textcolor{black}{↑2.38 }   &  55.64 &  \cellcolor{blue!5}\textcolor{black}{↑1.43 } &  52.16 &  \cellcolor{blue!5}\textcolor{black}{↑2.86 }  \\
& M. + \textit{w. SR, w CR.}              &  \textbf{69.03}  & \cellcolor{blue!5}\textbf{\textcolor{black}{↑3.99 }} &  \textbf{57.31} &  \cellcolor{blue!5}\textbf{\textcolor{black}{↑3.10 }} & \textbf{53.08} & \cellcolor{blue!5}\textbf{\textcolor{black}{↑3.78 }}  \\
    \bottomrule
\end{tabular}%
}
\end{center}
\label{tab:cr_ablation}
\vspace{-17pt}	
\end{table}

%%%%%%%%%%%%%%%%%%%%%%%%%%%%%%%%%%%%%%%%%%%%%%%%%%%%%%%%%%%%%%%%
\begin{table}[t]\footnotesize
\caption{Ablation study of multi-source curriculum learning order. The result obtained by the CLIP-VG model w/o extracting multi-level features. Annotations are the same as in \cref{tab:msa_execution}.}
\vspace{-12pt}
\begin{center}
\resizebox{0.90\columnwidth}{!}{%
\begin{tabular}{p{2.2cm} | p{1.6cm}<{\centering} | p{1.7cm}<{\centering} | p{1.2cm}<{\centering}}
    \toprule
    % \midrule
Curriculum Order & RefCOCO(val) & RefCOCO+(val) & ReferIt(val) \\
    \midrule
    \midrule
\textit{cap.-rel.-tmp.}  &  58.65 & 44.08 & 47.63  \\
\textit{rel.-tmp.-cap.}  &  60.87 & 45.32 & 49.79  \\
\textit{tmp.-cap.-rel.}  &  62.79 & 46.57 & 51.94  \\  \rowcolor{blue!5}
\textbf{\textit{tmp.-rel.-cap.}}  &  \textbf{62.86} & \textbf{48.40} & \textbf{53.85}  \\
    \bottomrule
\end{tabular}%
}
\end{center}
\label{tab:curriculum_order}
\vspace{-10pt}	
\end{table}

\begin{table}[t]\footnotesize
% \vspace{-5pt}
\caption{Comparison of results using different pre-trained backbones. The results are obtained on the testA split of the RefCOCO/+ dataset. \textit{w/o} represents ‘without’ using the SSA/MSA algorithm, while \textit{w.} represents ‘with’ using the SSA/MSA algorithm.}
\vspace{-17pt}
\begin{center}
\resizebox{1.0\columnwidth}{!}{%
\begin{tabular}{p{0.6cm} | p{1.8cm}<{\centering} | p{1.8cm}<{\centering} |  p{0.6cm}<{\centering} p{0.8cm}<{\centering}  |  p{0.6cm}<{\centering} p{0.8cm}<{\centering}}
    \toprule
    % \midrule
\multirow{2}{*}{Source} & \multirow{2}{*}{\makecell{Vision \\ Backbone}} & \multirow{2}{*}{\makecell{Language \\ Backbone}} &   \multicolumn{2}{c|}{RefCOCO} & \multicolumn{2}{c}{RefCOCO+} \\
& & & \textit{w/o} & \cellcolor{blue!5}\textit{\textbf{w.}} & \textit{w/o} & \cellcolor{blue!5}\textit{\textbf{w.}} \\
    \midrule    
    \midrule  
\multirow{4}{*}{\textit{\makecell{single- \\ source}}}
& ResNet-50     & BERT-base          & 57.91  & \cellcolor{blue!5}\textcolor{black}{\textbf{60.48 }} & 42.05 &  \cellcolor{blue!5}\textcolor{black}{\textbf{45.29 }}   \\
& ResNet-101    & BERT-base          & 58.40  & \cellcolor{blue!5}\textcolor{black}{\textbf{61.11 }} & 44.09 &  \cellcolor{blue!5}\textcolor{black}{\textbf{47.87 }}   \\
& CLIP-ViT-B/32 & CLIP-ViT-B/32      & 60.44  & \cellcolor{blue!5}\textcolor{black}{\textbf{64.12 }} & 50.93 &  \cellcolor{blue!5}\textcolor{black}{\textbf{53.89 }}   \\
& CLIP-ViT-B/16 & CLIP-ViT-B/16      & 61.92  & \cellcolor{blue!5}\textcolor{black}{\textbf{65.03 }} & 52.07 &  \cellcolor{blue!5}\textcolor{black}{\textbf{55.73 }}   \\
\midrule  \multirow{4}{*}{\textit{\makecell{multi- \\ source}}}
& ResNet-50     & BERT-base          & 58.29  & \cellcolor{blue!5}\textcolor{black}{\textbf{62.32 }} & 42.49 &  \cellcolor{blue!5}\textcolor{black}{\textbf{48.82 }}   \\
& ResNet-101    & BERT-base          & 59.10  & \cellcolor{blue!5}\textcolor{black}{\textbf{63.41 }} & 43.21 &  \cellcolor{blue!5}\textcolor{black}{\textbf{50.77 }}   \\
& CLIP-ViT-B/32 & CLIP-ViT-B/32      & 63.25  & \cellcolor{blue!5}\textcolor{black}{\textbf{67.72 }} & 52.07 &  \cellcolor{blue!5}\textcolor{black}{\textbf{55.48 }}   \\
& CLIP-ViT-B/16 & CLIP-ViT-B/16      & 65.04  & \cellcolor{blue!5}\textcolor{black}{\textbf{69.03 }} & 54.21 &  \cellcolor{blue!5}\textcolor{black}{\textbf{57.31 }}   \\
    \bottomrule
\end{tabular}%
}
\end{center}
\label{tab:backbone}
\vspace{-17pt}	
\end{table}

% \vspace{-1pt}
\vspace{+2pt}
\noindent\textbf{RefCOCO/RefCOCO+/RefCOCOg.} As shown in \cref{tab:SOTA1}, we provide results in both fully supervised and unsupervised settings. We compare our method with the existing SOTA unsupervised method Pseudo-Q\cite{jiang2022pseudo} in both single-source and multi-source scenarios. Although Pseudo-Q has greatly improved compared with previous works, our method can outperform Pseudo-Q on three datasets with a significant margin, improving by 6.78$\%$(testA), 10.67$\%$(testA), 7.37$\%$(test-u) in single-source and 14.65$\%$(testA), 14.87$\%$(testA), 11.39$\%$(test-u) in multi-source, respectively. Pseudo-labels can easily cause overfitting in a model. It can be seen that from single-source to multi-source, the performance of Pseudo-Q is degraded due to the influence of unreliable data (refer to \cref{tab:noise_proportion}), while our model avoids it. Furthermore, the results also outperform all of the weakly supervised methods, and the model is also competitive in the fully supervised setting.

It is worth noting that we did not compare MDETR\cite{kamath2021mdetr} in the fully supervised setting, as MDETR utilized a pre-training approach to retrain the backbone by using mixed grounding data from multiple datasets. Therefore, it would be unfair to compare its results with our work.

\vspace{+2pt}
\noindent\textbf{ReferItGame and Flickr30K Entities.} In \cref{tab:SOTA2}, our method achieves promising accuracy on the two datasets, which is higher than Pseudo-Q by 7.31$\%$ and 4.1$\%$ in single-source, and 9.77$\%$ and 9.85$\%$ in multi-source, and also outperforms all of the weakly supervised methods. 

% \vspace{-1pt}
\vspace{+2pt}
\noindent \textbf{Training/Inference Cost and Speed.} As shown in \cref{tab:cost}, we compare the current Transformer-based competitive models in terms of vision and language backbones, model parameters, training cost, and inference speed. The results are obtained on a single Nvidia 3090 GPU. The pre-trained backbones used by Pseudo-Q, TransVG, and MDETR are Resnet, BERT, and DETR, while QRNet uses Resnet, Swin Transformer, and BERT, and we only use CLIP-ViT-B/16. From the results, we can see that the existing fully supervised SOTA models (such as QRNet\cite{ye2022shifting}, MDETR\cite{kamath2021mdetr}) are particularly slow in both training and inference. Compared to QRNet, we updated \textbf{only 7.7$\%$} of its parameters and achieved impressive training and inference speedups, up to \textbf{26.84$\times$} and \textbf{7.41$\times$}, respectively, while also obtaining competitive results (\cref{tab:SOTA1} and \cref{tab:SOTA2}). Based on the reported results \cite{ho2023yoro}, our model is also state-of-the-art in terms of both speed and energy efficiency.

%%%%%%%%%%%%%%%%%%%%%%%%%%%%%%%%%%%%%%%%%%%%%%%%%%%%%%%%%%%%%%%%

\begin{figure*}[t]
\centering
  \includegraphics[width=0.9\linewidth]{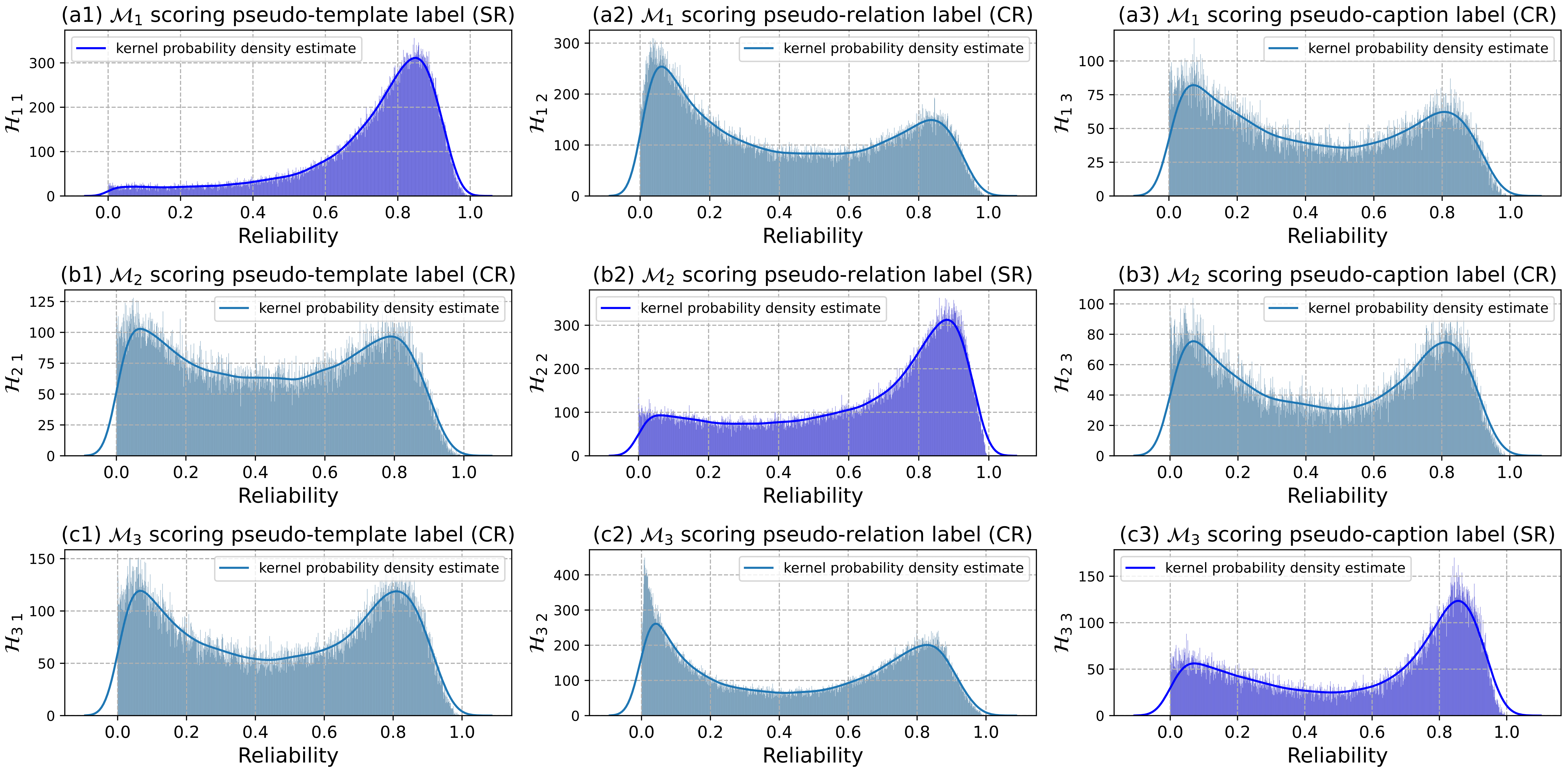}
  \vspace{-7pt}
  \caption{The complete Source-specific Reliability (SR, shown in \textcolor{blue}{blue color}) and Cross-source Reliability (CR, shown in \textcolor{teal}{teal color}) Histograms, which are formed by scoring the three sources of pseudo-language labels in the interval (0.0, 1.0] with different Measurers. $\mathcal{M}_1, \mathcal{M}_2,\mathcal{M}_3$ represent the Reliability Measurers learned from pseudo-template labels, pseudo-relation labels, and pseudo-caption labels, respectively. Different sources contain distinctive distributions due to specific quality and language taxonomy of pseudo-language labels (\ie, (a1)-(b2)-(c3)), and the different Reliability Measurer has divergent discrimination abilities on the same pseudo-label sources (\ie, (a1)-(b1)-(c1)).}
  \label{fig:sup_reliability}
  \vspace{-5pt}
\end{figure*}

\vspace{-5pt}
\subsection{Ablation Study}
\label{4.3ablation}
\vspace{-2pt}

\vspace{+2pt}
\noindent \textbf{Ablation of SSA and MSA Algorithms.} \cref{tab:SOTA1} and \cref{tab:SOTA2} demonstrate the performance improvements achieved by utilizing the SSA and MSA algorithms. It can be seen that the performance gains brought by the SSA and MSA algorithms are 3.11$\%$(testA), 3.66$\%$(testA), 3.34$\%$(test-u) in single source, and 3.89$\%$(testA), 3.10$\%$(testA), 3.07$\%$(test-u) in multi-source, respectively. \textbf{Notably,} our work's improvements in the multi-source scenario are primarily attributed to the proposed MSA algorithm rather than utilizing more sources of pseudo-labels. \cref{tab:msa_execution} demonstrates the performance gains achieved by each step on both SSA and MSA. It is evident that stacking multi-source pseudo-labels leads to a decline in performance. The results indicate that the performance steadily increased with the implementation of the MSA algorithm, ultimately resulting in significant improvement. Our method exhibits strong superiority in the multi-source scenario.

\vspace{+2pt}
\noindent \textbf{Ablation of Cross-source Reliability (CR).} Only Source-specific Reliability (SR) is utilized in the case of single-source, while both SR and CR are utilized in the case of multi-source. In the single-source scenario, the model learns from the specific source and captures its primary characteristics in pseudo data. We can utilize SR to select more reliable data and reduce the impact of unreliable pseudo-labels. In the multi-source scenario, the model learned from the current source is easily biased from the ideal model due to discrepancies between the pseudo-label and the ground-truth label, which may affect the effectiveness of data selection. By further considering CR, we can use models learned from other sources to guide the pseudo-label selection in the current source and sample more generalized pseudo triplet data. \cref{tab:cr_ablation} shows the ablation results of CR in the multi-source scenario, indicating that it contributes to the performance gains by 1.51$\%$, 1.67$\%$, and 0.92$\%$, respectively.

\vspace{+2pt}
\noindent \textbf{Ablation of Multi-source Curriculum Learning Order.} In \cref{3.5MSA}, we propose the source-level complexity metric, \ie, \textit{the average number of entities per expression}. The complexity values of different pseudo-labels calculated in the experiment are as follows: \textit{tmp.:}$1.1562$, \textit{rel.:}$1.8882$, \textit{cap.:}$3.1961$. Thus, MSA is performed in the order of \textit{tmp.-rel.-cap.}. \cref{tab:curriculum_order} shows the results (val split) when changing the learning order, which verifies the effectiveness of our proposed curriculum order.

\vspace{+2pt}
\noindent 
\textbf{Generality of SSA and MSA Algorithms.} Our main experimental results are achieved with CLIP-ViT-B/16, but our proposed algorithms are general and not limited to CLIP. \cref{tab:backbone} shows the results obtained by using different backbones. It can be seen that both the SSA and MSA algorithms can improve the results of the original model by about 3+$\%$.

\vspace{-5pt}
\subsection{Further Remarks}
\label{4.4remark}
\vspace{-3pt}

\vspace{+2pt}
\noindent\textbf{Visualization of Reliability Histogram.} \cref{fig:sup_reliability} presents the histograms of Single-Source Reliability (SR) and Cross-source Reliability (CR) for pseudo-language labels in the range of (0.0, 1.0] with 1000 bins, where each bin represents the number of samples. The figure illustrates that different sources exhibit distinct distributions due to their specific quality and language taxonomy of pseudo-language labels (\eg, \cref{fig:sup_reliability}\textcolor{red}{-(a1)-(b2)-(c3)}), while different reliability measures have varying discrimination abilities on the same source (\eg, \cref{fig:sup_reliability}\textcolor{red}{-(a1)-(b1)-(c1)}). This provides an explanation for the performance gains of our approach.

%%%%%%%%%%%%%%%%%%%%%%%%%%%%%%%%%%%%%%%%%%%%%%%%%%%%%%%%%%%%%%%%
\begin{figure*}[t]
\centering
  \includegraphics[width=0.9\linewidth]{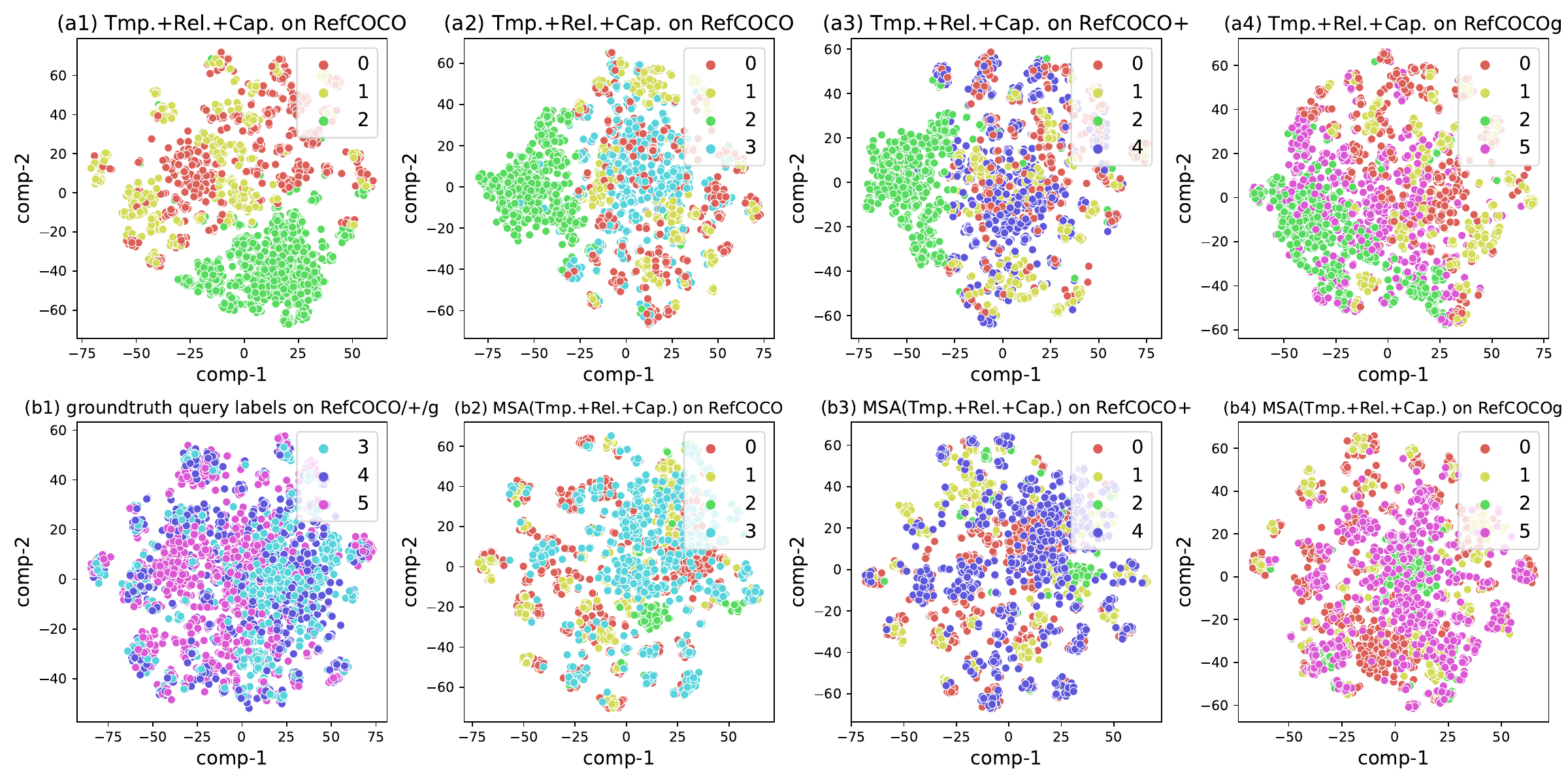}
  \vspace{-10pt}
  \caption{The CLIP text feature of the pseudo-language labels and the ground-truth query labels on RefCOCO/+/g datasets are visualized by using t-SNE. The figure shows the comparison before and after MSA execution, and the generalization results of pseudo-language labels on the ground-truth query labels. The legend: 0-pseudo-template label, 1-pseudo-relation label, 2-pseudo-caption label, 3,4,5-the ground-truth query labels on RefCOCO/+/g val split. (a1) shows the distribution discrepancy of semantic features on language taxonomy for the pseudo-language labels on the RefCOCO dataset, while (b1) shows this distribution discrepancy for the ground-truth query labels on the RefCOCO/+/g dataset validation split. The comparison is (a2)-(b2), (a3)-(b3), (a4)-(b4). The feature distribution of pseudo-language labels after the execution of the MSA algorithm basically fits the distribution of the ground-truth query labels (\ie, (b2), (b3), (b4)). This figure shows one of the reasons for the performance gain of MSA algorithm, namely, the feature generalization for language taxonomy.}
  \label{fig:t-sne}
  % \vspace{\figmargin}
  \vspace{-10pt}
\end{figure*}

%%%%%%%%%%%%%%%%%%%%%%%%%%%%%%%%%%%%%%%%%%%%%%%%%%%%%%%%%%%%%%%%
\begin{figure}[t]
\centering
  \includegraphics[width=0.95\linewidth]{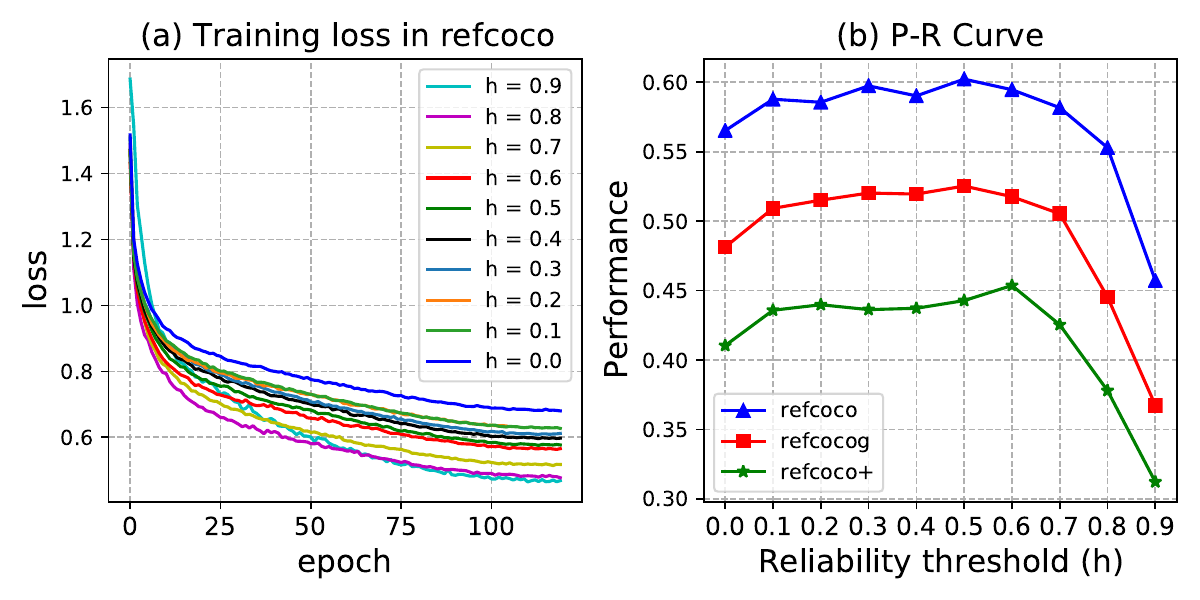}
  \vspace{-7pt}
  \caption{The result with reliability threshold $\mathscr{h}$ from 0.9 to 0 during the execution of the SSA algorithm on the RefCOCO/+/g datasets (val split). (a) Convergence curve on the RefCOCO dataset, and (b) P-R curve.}
  \label{fig:p-r_curve}
  \vspace{-9pt}
\end{figure}

%%%%%%%%%%%%%%%%%%%%%%%%%%%%%%%%%%%%%%%%%%%%%%%%%%%%%%%%%%%%%%%%
\begin{table}[t]\footnotesize
\caption{The proportion of the RefCOCO dataset's most unreliable data ($\mathscr{r}=0$) in the three pseudo-label sources, which is measured by Source-specific Reliability (SR).}
\vspace{-12pt}
\begin{center}
\resizebox{0.99\columnwidth}{!}{%
\begin{tabular}{p{3.4cm} | p{1.3cm}<{\centering} | p{1.3cm}<{\centering} | p{1.3cm}<{\centering}}
    \toprule
    % \midrule
 Item & \textit{tmp.} label & \textit{rel.} label & \textit{cap.} label \\
    \midrule
    \midrule
expression num        &  95982 & 156897 & 60797  \\
num of most unreliable labels   &  5296 & 33473 & 12330  \\  \rowcolor{blue!5}
proportion    &  5.52$\%$ & 21.33$\%$ & 20.28$\%$  \\ 
    \bottomrule
\end{tabular}%
}
\end{center}
\label{tab:noise_proportion}
\vspace{-12pt}	
\end{table}

\begin{figure}[t]
\centering
   \includegraphics[width=0.95\linewidth]{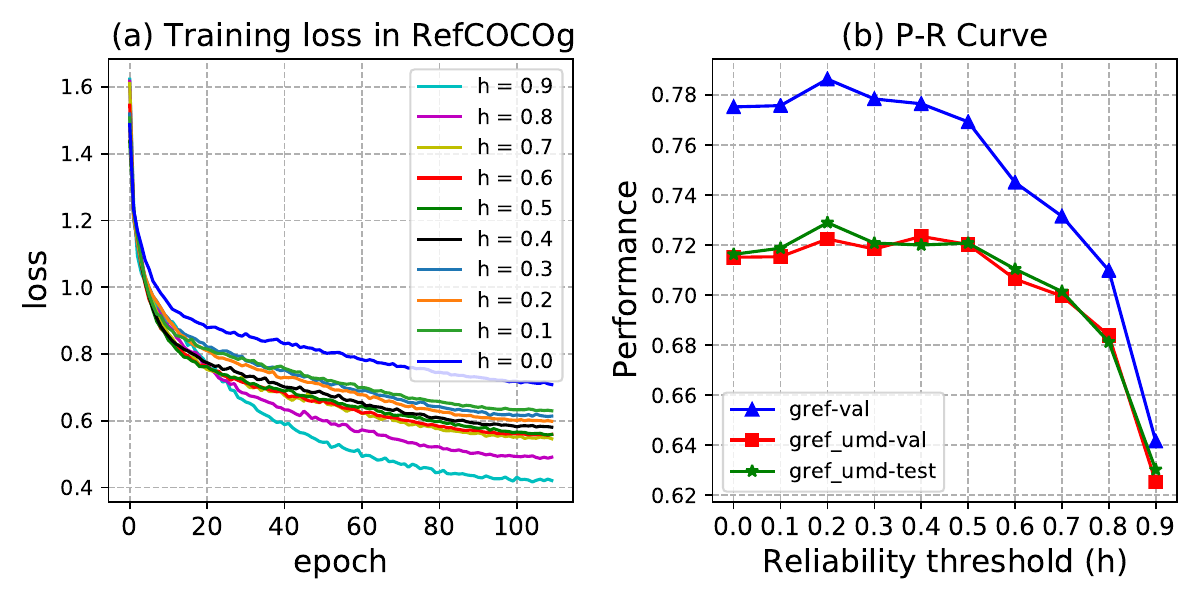}
    \vspace{-7pt}
   \caption{The result with reliability threshold $\mathscr{h}$ from 0.9 to 0 during the execution of SSA algorithm on the three splits (gref-val/gref-umd-val/gref-umd-test) of RefCOCOg dataset in the fully supervised setting. (a) Convergence curve, and (b) P-R curve.}
   \label{fig:sup_covg}
    \vspace{-10pt}
\end{figure}

%%%%%%%%%%%%%%%%%%%%%%%%%%%%%%%%%%%%%%%%%%%%%%%%%%%%%%%%%%%%%%%%%%%%%
\begin{table}[t]\footnotesize
% \vspace{\figmargin}
\caption{Performance improvement of Single-source Self-paced Curriculum Adapting (SSA) algorithm in fully supervised setting on RefCOCO/+/g datasets. \textit{w/o} represents ‘without’, \textit{w.} represents ‘with’.}
\vspace{-20pt}
\begin{center}
\resizebox{1.0\columnwidth}{!}{%
\begin{tabular}{p{2.2cm} |  p{0.7cm}<{\centering} p{0.6cm}<{\centering} p{0.7cm}<{\centering} | p{0.7cm}<{\centering} p{0.6cm}<{\centering} p{0.7cm}<{\centering} | p{0.7cm}<{\centering} p{0.6cm}<{\centering} p{0.7cm}<{\centering}}
    \toprule
\multirow{2}{*}{Method} &  \multicolumn{3}{c|}{RefCOCO} & \multicolumn{3}{c|}{RefCOCO+} & \multicolumn{3}{c}{RefCOCOg} \\
 &  val & testA & testB & val & testA & testB & val-g & val-u & test-u \\
    \midrule
    \midrule
TransVG \cite{deng2021transvg} &  80.49  & 83.28  & 75.24  & 66.39  & 70.55  & 57.66  & 66.35  & 67.93  & 67.44  \\  \rowcolor{blue!5}
TransVG \textit{\textbf{w.}} SSA                 &  \textbf{81.47}  & \textbf{83.87}  & \textbf{75.74}  & \textbf{66.66}  & \textbf{71.95}  & \textbf{57.71}  & \textbf{68.68}  & \textbf{68.48}  & \textbf{68.51}  \\
    \midrule
QRNet \cite{ye2022shifting}    &  84.01  & 85.85  & 82.34  & 72.94  & 76.17  & 63.81  & 71.89  & 73.03  & 72.52  \\  \rowcolor{blue!5}
QRNet \textit{\textbf{w.}} SSA                   &  \textbf{84.21}  & \textbf{86.62}  & \textbf{82.38}  & \textbf{73.16}  & \textbf{76.56}  & \textbf{64.86}  & \textbf{73.38}  & \textbf{74.11}  & \textbf{72.61}  \\ 
    \midrule
CLIP-VG \textit{w/o} SSA                &  84.11  & 87.63  & 78.13  & 68.45  & 77.14  & 56.44  & 72.43  & 72.08  & 72.02  \\  \rowcolor{blue!5}
CLIP-VG \textit{\textbf{w.}} SSA                 &  \textbf{84.29}  & \textbf{87.76}  & \textbf{78.43}  & \textbf{69.55}  & \textbf{77.33}  & \textbf{57.62}  & \textbf{72.64}  & \textbf{73.18}  & \textbf{72.54}  \\
    \bottomrule\end{tabular}%
}
\end{center}
\label{tab:ssa_perf}
\vspace{-15pt}	
\end{table}

\begin{figure*}[!h]
\centering
  \includegraphics[width=0.93\linewidth]{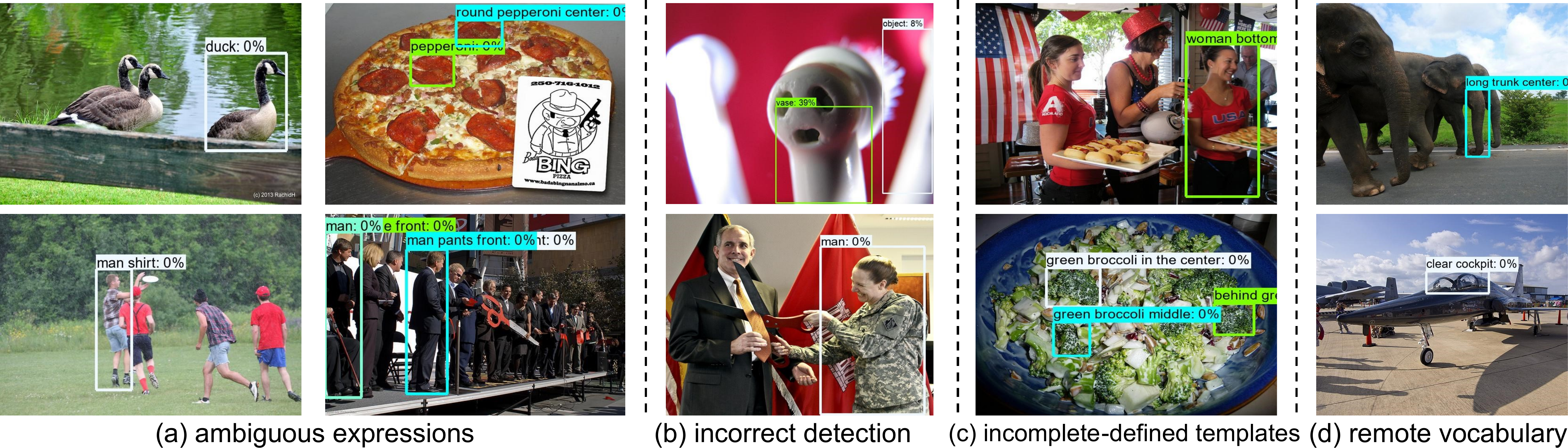}
  \vspace{-7pt}
  \caption{Examples of most unreliable data in pseudo-template labels. The percentage following the label indicates the Reliability value, as do \cref{fig:noise_data_rel,fig:noise_data_cap}. (Best view in color and zoom in.)}  \label{fig:noise_data_tmp}
\vspace{-5pt}
\end{figure*}
\begin{figure*}[!h]

\centering
  \includegraphics[width=0.93\linewidth]{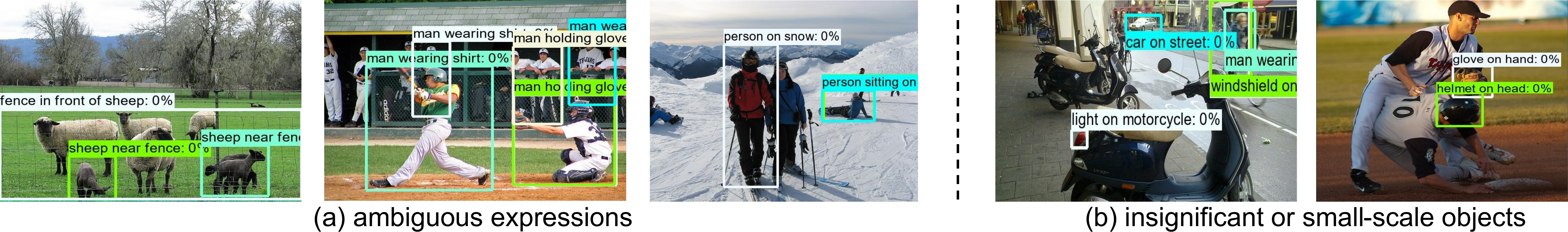}
  \vspace{-7pt}
  \caption{Examples of most unreliable data in pseudo-relation labels. (Best view in color and zoom in.)}
  \label{fig:noise_data_rel}
\vspace{-5pt}
\end{figure*}

\begin{figure*}[!h]
\centering
  \includegraphics[width=0.93\linewidth]{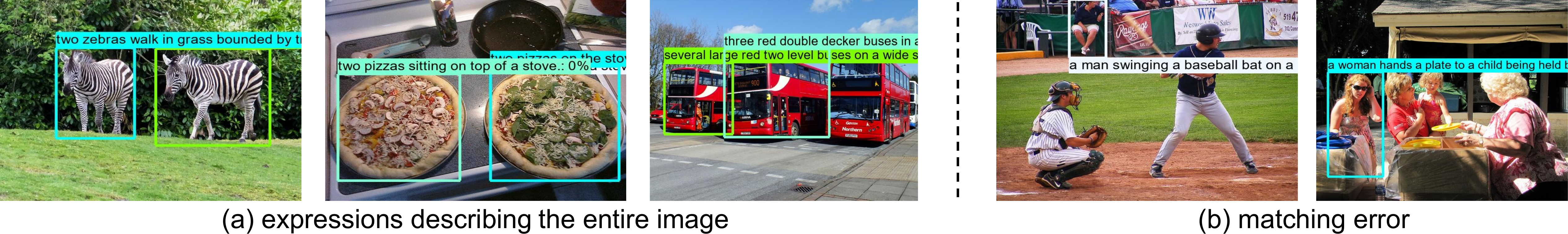}
  \vspace{-7pt}
  \caption{Examples of most unreliable data in pseudo-caption labels. (Best view in color and zoom in.)}
  \label{fig:noise_data_cap}
\vspace{-10pt}
\end{figure*}

\vspace{+2pt}
\noindent\textbf{Visualization of MSA in Generalization Ability.}
% \noindent\textbf{Discrepancy in Feature Distribution.} 
As shown in \cref{fig:t-sne}, we use t-SNE to visualize the CLIP text feature of the pseudo-language labels and the ground-truth query labels on RefCOCO/+/g datasets. \cref{fig:t-sne}\textcolor{red}{-(a1)} is the feature of three pseudo-labels on the RefCOCO dataset, and \cref{fig:t-sne}\textcolor{red}{-(b1)} is the feature of the ground-truth query labels on RefCOCO/+/g validation split, which respectively shows the feature distribution discrepancy among the three pseudo-label sources and the three ground-truth query labels. \cref{fig:t-sne}\textcolor{red}{-(a2)-(a4)} and \cref{fig:t-sne}\textcolor{red}{-(b2)-(b4)} are the feature distribution comparison of three pseudo-label sources and the ground-truth query labels before and after using MSA on RefCOCO/+/g datasets, respectively. 
Before the execution of MSA, the distribution of the pseudo-language labels and the ground-truth query labels is quite different, but after the execution of MSA, the distribution discrepancy significantly becomes smaller. This shows that MSA can effectively select pseudo-labels that are more reliable or closer to the distribution of ground-truth query labels.

\vspace{+2pt}
\noindent\textbf{Performance-Reliability (P-R) Curve and Convergence.} During the greedy sample selection in SSA and MSA algorithms, we sample the pseudo-labels that have reliability values belonging to the interval $[\mathscr{h}, 1.0]$ of the Reliability Histogram $\mathcal{H}_{i^*j^*}$, and then add the selected samples to the subset $\mathcal{D}_\chi$ to construct a temporary subset, where $\mathcal{D}_\chi$ is the whole set of selected pseudo samples before current SSA or MSA step. We draw the Performance Reliability (P-R) curve to reflect the performance of the model trained by the temporary subset obtained with different values of the reliability threshold $\mathscr{h}$. The greedy sample selection aims to find the reliability threshold corresponding to a local extreme point on the P-R curve to balance the reliable and unreliable pseudo-labels. 

\cref{fig:p-r_curve} illustrates the training loss and performance curve during the execution of greedy sample selection in SSA with the Reliability threshold from 0.9 to 0. In \cref{fig:p-r_curve}\textcolor{red}{-(a)}, the higher value of $\mathscr{h}$ leads to faster model convergence and the smaller converged loss. For the P-R curve in \cref{fig:p-r_curve}\textcolor{red}{-(b)}, the model achieves performance saturation in range of $[0.4, 0.6]$, which is the reason for $\mathscr{h}_0$ set $0$.

% \vspace{+2pt}
\noindent \textbf{Analysis of Most Unreliable Data.} The most unreliable data is represented by $\mathscr{r}=0$. As shown in \cref{fig:p-r_curve}\textcolor{red}{-(b)}, when $\mathscr{h}$ approaches 0, the accuracy decreases significantly. Our algorithm filters out the most unreliable data as demonstrated in \cref{tab:noise_proportion}, thus preventing its harmful effects.

\subsection{Application of SSA in Fully Supervised Setting}
\label{subsec:fully_exp}

\vspace{+2pt}
\noindent\textbf{Performance of SSA in Fully Supervised VG.} We use CLIP-VG, TransVG\cite{deng2021transvg}, and QRNet\cite{ye2022shifting} as baseline models to verify the effectiveness of Single-source Self-paced Adapting algorithm (SSA) under the fully supervised setting. As shown in \cref{tab:ssa_perf}, the SSA can further improve the original model's performance in most cases.

\vspace{+2pt}
\noindent\textbf{Convergence Analysis.}
\cref{fig:sup_covg}\textcolor{red}{-(a)} illustrates the training loss curve during the execution of SSA on refCOCOg dataset with Reliability threshold $\mathscr{h}$ ranging from 0.9 to 0, where a higher value of $\mathscr{h}$ leads to faster model convergence and smaller converged loss.

\vspace{+2pt}
\noindent\textbf{Performance-Reliability (P-R) Curve.}
As the P-R curve in \cref{fig:sup_covg}\textcolor{red}{-(b)} shows, the model achieves performance saturation in the range of $[0.05, 0.25]$, which provides a prior for the SSA algorithm in the fully supervised setting, that is $\mathscr{h}_0$ should be set to 0.2. It should be noted that due to the high quality of the manual annotation, the performance saturation point (\textit{i.e.}, 0.2) of the reliability threshold is smaller than that of the pseudo-labels (\textit{i.e.}, 0.5). The accuracy will suffer a decrease when the Reliability threshold gets closer to 0, which to some extent reflects that there is still a certain proportion of unreliable samples in the manually labeled annotations \cite{wang2021survey}. This indicates the boundaries of performance on RefCOCO/+/g datasets.

\subsection{Qualitative Analysis of Unreliable Pseudo-language Labels}
\label{sec:qualitative}

In this section, we study the most unreliable pseudo-language labels that have been successfully filtered and eliminated by our SSA and MSA algorithms, while also providing visual representations of these most unreliable data. 

As shown in \cref{tab:noise_proportion}, a large number of pseudo-labels are concentrated at Reliability $\mathscr{r}=0$, which significantly reduces the model's performance (as the P-R curve shown in \cref{fig:p-r_curve}). When $\mathscr{r}=0$, it means that the referred region cannot be localized, which seriously hinders the model from acquiring correct knowledge. By using SSA and MSA to eliminate these unreliable data points, both pseudo-labels and manually annotated data can further improve the model's performance. The specific most unreliable pseudo-language labels ($\mathscr{r}=0$) are shown in \cref{fig:noise_data_tmp,fig:noise_data_rel,fig:noise_data_cap}.

In the pseudo-template label (\cref{fig:noise_data_tmp}), we roughly divide the unreliable data into four categories: (a) ambiguous expressions, \ie, lack of uniqueness; (b) wrong labels caused by incorrect detection results; (c) incomplete prior information (for example, the spatial relationship defined in Pseudo-Q, \eg, ‘front’, ‘middle’, ‘bottom’ are not accurate); (d) other issues, such as remote vocabulary, insignificant or small-scale objects, \textit{etc.}

In the pseudo-relation label (\cref{fig:noise_data_rel}), we roughly divide the unreliable data into (a) ambiguous expressions, and (b) insignificant or small-scale objects.

In the pseudo-caption label (\cref{fig:noise_data_cap}), we roughly divide the unreliable data into (a) the pseudo-language labels describing the entire image, and (b) mismatches between bounding boxes and captions.

Among the various types of unreliable pseudo-language labels, referring to ambiguity is more frequent, particularly in images with similar classification objects. If future research aims to further enhance model performance, addressing ambiguity is a critical issue.

%%%%%%%%%%%%%%%%%%%%%%%%%%%%%%%%%%%%%%%%%%%%%%%%%%%%%%%%%%%%%%%%%%%%%%%%%
%------------------------------------------------------------------------
%------------------------------------------------------------------------
%------------------------------------------------------------------------
%------------------------------------------------------------------------
\section{Discussion}
\label{sec:discussion}
% \vspace{+2pt}

\noindent\textbf{Explanation of Performance Gains.} The key to completing the grounding task lies in comprehending the correspondence between language expression and image regions. Our approach introduces pseudo-language labels and pseudo-label quality measurement for unsupervised settings. The SSA and MSA algorithms achieve an optimal balance between reliable and unreliable pseudo-labels, resulting in more stable learning of the CLIP-based visual grounding model, which significantly improves the model's generalization.

\vspace{+2pt}
\noindent\textbf{Limitations.} We have introduced three types of pseudo-labels, but their quality remains low. In order to strike a balance between reliable and unreliable labels, we exclude the latter and do not further utilize them, even though they still contain valuable information. Furthermore, the greedy sample selection strategy employed in both SSA and MSA represents a trade-off between training cost and optimal solution. These can be further explored in future research.

%%%%%%%%%%%%%%%%%%%%%%%%%%%%%%%%%%%%%%%%%%%%%%%%%%%%%%%%%%%%%%%%%%%%%%%%%
%------------------------------------------------------------------------
%------------------------------------------------------------------------
%------------------------------------------------------------------------
%------------------------------------------------------------------------
\section{Conclusion}
\label{5.0conclusion}
% \vspace{-5pt}

In this paper, we propose a novel CLIP-VG method that enables the unsupervised transfer of CLIP to the grounding task by incorporating pseudo-language labels. This is the first attempt to apply the concept of self-paced curriculum adapting to visual grounding. As downstream vision and language contexts continue to evolve, multiple sources of pseudo-labeling are likely to become a future trend. Our proposed multi-source pseudo-language labels and the curriculum adapting method offer a fresh perspective for future research. The idea of our approach is simple yet effective, and it may be used as a plug-in in various cross-modal pseudo-labeling tasks in the future.

%%%%%%%%%%%%%%%%%%%%%%%%%%%%%%%%%%%%%%%%%%%%%%%%%%%%%%%%%%%%%%%%%%%%%%%%%%%
%----------------------
%----------------------
%----------------------
% \section*{Acknowledgments}
% This should be a simple paragraph before the References to thank those individuals and institutions who have supported your work on this article.

%%%%%%%%%%%%%%%%%%%%%%%%%%%%%%%%%%%%%%%%%%%%%%%%%%%%%%%%%%%%%%%%%%%%%%%%%%%
%----------------------
%----------------------
%----------------------
\bibliographystyle{IEEEtran}
\bibliography{ref}

\begin{thebibliography}{10}
\providecommand{\url}[1]{#1}
\csname url@samestyle\endcsname
\providecommand{\newblock}{\relax}
\providecommand{\bibinfo}[2]{#2}
\providecommand{\BIBentrySTDinterwordspacing}{\spaceskip=0pt\relax}
\providecommand{\BIBentryALTinterwordstretchfactor}{4}
\providecommand{\BIBentryALTinterwordspacing}{\spaceskip=\fontdimen2\font plus
\BIBentryALTinterwordstretchfactor\fontdimen3\font minus \fontdimen4\font\relax}
\providecommand{\BIBforeignlanguage}[2]{{%
\expandafter\ifx\csname l@#1\endcsname\relax
\typeout{** WARNING: IEEEtran.bst: No hyphenation pattern has been}%
\typeout{** loaded for the language `#1'. Using the pattern for}%
\typeout{** the default language instead.}%
\else
\language=\csname l@#1\endcsname
\fi
#2}}
\providecommand{\BIBdecl}{\relax}
\BIBdecl

\bibitem{qiao2020referring}
Y.~Qiao, C.~Deng, and Q.~Wu, ``Referring expression comprehension: A survey of methods and datasets,'' \emph{IEEE Transactions on Multimedia}, vol.~23, pp. 4426--4440, 2020.

\bibitem{mao2016generation}
J.~Mao, J.~Huang, A.~Toshev, O.~Camburu, A.~L. Yuille, and K.~Murphy, ``Generation and comprehension of unambiguous object descriptions,'' in \emph{Proceedings of the IEEE/CVF Conference on Computer Vision and Pattern Recognition}, 2016.

\bibitem{yu2016modeling}
L.~Yu, P.~Poirson, S.~Yang, A.~C. Berg, and T.~L. Berg, ``Modeling context in referring expressions,'' in \emph{Computer Vision--ECCV 2016: 14th European Conference, Amsterdam, The Netherlands, October 11-14, 2016, Proceedings, Part II 14}.\hskip 1em plus 0.5em minus 0.4em\relax Springer, 2016, pp. 69--85.

\bibitem{hu2016natural}
R.~Hu, H.~Xu, M.~Rohrbach, J.~Feng, K.~Saenko, and T.~Darrell, ``Natural language object retrieval,'' in \emph{Proceedings of the IEEE/CVF Conference on Computer Vision and Pattern Recognition}, 2016.

\bibitem{deng2021transvg}
J.~Deng, Z.~Yang, T.~Chen, W.~Zhou, and H.~Li, ``Transvg: End-to-end visual grounding with transformers,'' in \emph{Proceedings of the IEEE/CVF International Conference on Computer Vision}, 2021.

\bibitem{antol2015vqa}
S.~Antol, A.~Agrawal, J.~Lu, M.~Mitchell, D.~Batra, C.~L. Zitnick, and D.~Parikh, ``Vqa: Visual question answering,'' in \emph{Proceedings of the IEEE/CVF International Conference on Computer Vision}, 2015.

\bibitem{anderson2018vision}
P.~Anderson, Q.~Wu, D.~Teney, J.~Bruce, M.~Johnson, N.~S{\"u}nderhauf, I.~Reid, S.~Gould, and A.~Van Den~Hengel, ``Vision-and-language navigation: Interpreting visually-grounded navigation instructions in real environments,'' in \emph{Proceedings of the IEEE conference on computer vision and pattern recognition}, 2018, pp. 3674--3683.

\bibitem{chen2018real}
X.~Chen, L.~Ma, J.~Chen, Z.~Jie, W.~Liu, and J.~Luo, ``Real-time referring expression comprehension by single-stage grounding network,'' \emph{arXiv preprint arXiv:1812.03426}, 2018.

\bibitem{liao2020real}
Y.~Liao, S.~Liu, G.~Li, F.~Wang, Y.~Chen, C.~Qian, and B.~Li, ``A real-time cross-modality correlation filtering method for referring expression comprehension,'' in \emph{Proceedings of the IEEE/CVF Conference on Computer Vision and Pattern Recognition}, 2020.

\bibitem{hong2019learning}
R.~Hong, D.~Liu, X.~Mo, X.~He, and H.~Zhang, ``Learning to compose and reason with language tree structures for visual grounding,'' \emph{IEEE transactions on pattern analysis and machine intelligence}, vol.~44, no.~2, pp. 684--696, 2019.

\bibitem{hu2017modeling}
R.~Hu, M.~Rohrbach, J.~Andreas, T.~Darrell, and K.~Saenko, ``Modeling relationships in referential expressions with compositional modular networks,'' in \emph{Proceedings of the IEEE/CVF Conference on Computer Vision and Pattern Recognition}, 2017.

\bibitem{liu2019learning}
D.~Liu, H.~Zhang, F.~Wu, and Z.-J. Zha, ``Learning to assemble neural module tree networks for visual grounding,'' in \emph{Proceedings of the IEEE/CVF International Conference on Computer Vision}, 2019.

\bibitem{sun2021cycle}
M.~Sun, J.~Xiao, E.~G. Lim, and Y.~Zhao, ``Cycle-free weakly referring expression grounding with self-paced learning,'' \emph{IEEE Transactions on Multimedia}, 2021.

\bibitem{wang2021weakly}
Y.~Wang, J.~Deng, W.~Zhou, and H.~Li, ``Weakly supervised temporal adjacent network for language grounding,'' \emph{IEEE Transactions on Multimedia}, vol.~24, pp. 3276--3286, 2021.

\bibitem{chen2018knowledge}
K.~Chen, J.~Gao, and R.~Nevatia, ``Knowledge aided consistency for weakly supervised phrase grounding,'' in \emph{Proceedings of the IEEE/CVF Conference on Computer Vision and Pattern Recognition}, 2018.

\bibitem{datta2019align2ground}
S.~Datta, K.~Sikka, A.~Roy, K.~Ahuja, D.~Parikh, and A.~Divakaran, ``Align2ground: Weakly supervised phrase grounding guided by image-caption alignment,'' in \emph{Proceedings of the IEEE/CVF International Conference on Computer Vision}, 2019.

\bibitem{gupta2020contrastive}
T.~Gupta, A.~Vahdat, G.~Chechik, X.~Yang, J.~Kautz, and D.~Hoiem, ``Contrastive learning for weakly supervised phrase grounding,'' in \emph{Computer Vision--ECCV 2020: 16th European Conference, Glasgow, UK, August 23--28, 2020, Proceedings, Part III}.\hskip 1em plus 0.5em minus 0.4em\relax Springer, 2020, pp. 752--768.

\bibitem{liu2021relation}
Y.~Liu, B.~Wan, L.~Ma, and X.~He, ``Relation-aware instance refinement for weakly supervised visual grounding,'' in \emph{Proceedings of the IEEE/CVF Conference on Computer Vision and Pattern Recognition}, 2021.

\bibitem{wang2021improving}
L.~Wang, J.~Huang, Y.~Li, K.~Xu, Z.~Yang, and D.~Yu, ``Improving weakly supervised visual grounding by contrastive knowledge distillation,'' in \emph{Proceedings of the IEEE/CVF Conference on Computer Vision and Pattern Recognition}, 2021.

\bibitem{yeh2018unsupervised}
R.~A. Yeh, M.~N. Do, and A.~G. Schwing, ``Unsupervised textual grounding: Linking words to image concepts,'' in \emph{Proceedings of the IEEE/CVF Conference on Computer Vision and Pattern Recognition}, 2018.

\bibitem{wang2019phrase}
J.~Wang and L.~Specia, ``Phrase localization without paired training examples,'' in \emph{Proceedings of the IEEE/CVF International Conference on Computer Vision}, 2019.

\bibitem{shi2022unpaired}
H.~Shi, M.~Hayat, and J.~Cai, ``Unpaired referring expression grounding via bidirectional cross-modal matching,'' \emph{arXiv preprint arXiv:2201.06686}, 2022.

\bibitem{jiang2022pseudo}
H.~Jiang, Y.~Lin, D.~Han, S.~Song, and G.~Huang, ``Pseudo-q: Generating pseudo language queries for visual grounding,'' in \emph{Proceedings of the IEEE/CVF Conference on Computer Vision and Pattern Recognition}, 2022, pp. 15\,513--15\,523.

\bibitem{radford2021learning}
A.~Radford, J.~W. Kim, C.~Hallacy, A.~Ramesh, G.~Goh, S.~Agarwal, G.~Sastry, A.~Askell, P.~Mishkin, J.~Clark \emph{et~al.}, ``Learning transferable visual models from natural language supervision,'' in \emph{International conference on machine learning}.\hskip 1em plus 0.5em minus 0.4em\relax PMLR, 2021, pp. 8748--8763.

\bibitem{wang2022image}
W.~Wang, H.~Bao, L.~Dong, J.~Bjorck, Z.~Peng, Q.~Liu, K.~Aggarwal, O.~K. Mohammed, S.~Singhal, S.~Som \emph{et~al.}, ``Image as a foreign language: Beit pretraining for all vision and vision-language tasks,'' \emph{arXiv preprint arXiv:2208.10442}, 2022.

\bibitem{kazemzadeh2014referitgame}
S.~Kazemzadeh, V.~Ordonez, M.~Matten, and T.~Berg, ``Referitgame: Referring to objects in photographs of natural scenes,'' in \emph{Proceedings of the 2014 conference on empirical methods in natural language processing (EMNLP)}, 2014, pp. 787--798.

\bibitem{plummer2015flickr30k}
B.~A. Plummer, L.~Wang, C.~M. Cervantes, J.~C. Caicedo, J.~Hockenmaier, and S.~Lazebnik, ``Flickr30k entities: Collecting region-to-phrase correspondences for richer image-to-sentence models,'' in \emph{Proceedings of the IEEE/CVF International Conference on Computer Vision}, 2015.

\bibitem{ye2022shifting}
J.~Ye, J.~Tian, M.~Yan, X.~Yang, X.~Wang, J.~Zhang, L.~He, and X.~Lin, ``Shifting more attention to visual backbone: Query-modulated refinement networks for end-to-end visual grounding,'' in \emph{Proceedings of the IEEE/CVF Conference on Computer Vision and Pattern Recognition}, 2022, pp. 15\,502--15\,512.

\bibitem{ho2023yoro}
C.-H. Ho, S.~Appalaraju, B.~Jasani, R.~Manmatha, and N.~Vasconcelos, ``Yoro-lightweight end to end visual grounding,'' in \emph{Computer Vision--ECCV 2022 Workshops: Tel Aviv, Israel, October 23--27, 2022, Proceedings, Part VIII}.\hskip 1em plus 0.5em minus 0.4em\relax Springer, 2023, pp. 3--23.

\bibitem{vaswani2017attention}
A.~Vaswani, N.~Shazeer, N.~Parmar, J.~Uszkoreit, L.~Jones, A.~N. Gomez, {\L}.~Kaiser, and I.~Polosukhin, ``Attention is all you need,'' \emph{Advances in neural information processing systems}, vol.~30, 2017.

\bibitem{dosovitskiy2020image}
A.~Dosovitskiy, L.~Beyer, A.~Kolesnikov, D.~Weissenborn, X.~Zhai, T.~Unterthiner, M.~Dehghani, M.~Minderer, G.~Heigold, S.~Gelly \emph{et~al.}, ``An image is worth 16x16 words: Transformers for image recognition at scale,'' \emph{arXiv preprint arXiv:2010.11929}, 2020.

\bibitem{carion2020end}
N.~Carion, F.~Massa, G.~Synnaeve, N.~Usunier, A.~Kirillov, and S.~Zagoruyko, ``End-to-end object detection with transformers,'' in \emph{Computer Vision--ECCV 2020: 16th European Conference, Glasgow, UK, August 23--28, 2020, Proceedings, Part I 16}.\hskip 1em plus 0.5em minus 0.4em\relax Springer, 2020, pp. 213--229.

\bibitem{sadhu2019zero}
A.~Sadhu, K.~Chen, and R.~Nevatia, ``Zero-shot grounding of objects from natural language queries,'' in \emph{Proceedings of the IEEE/CVF International Conference on Computer Vision}, 2019.

\bibitem{yang2020improving}
Z.~Yang, T.~Chen, L.~Wang, and J.~Luo, ``Improving one-stage visual grounding by recursive sub-query construction,'' in \emph{Computer Vision--ECCV 2020: 16th European Conference, Glasgow, UK, August 23--28, 2020, Proceedings, Part XIV 16}.\hskip 1em plus 0.5em minus 0.4em\relax Springer, 2020, pp. 387--404.

\bibitem{yu2018mattnet}
L.~Yu, Z.~Lin, X.~Shen, J.~Yang, X.~Lu, M.~Bansal, and T.~L. Berg, ``Mattnet: Modular attention network for referring expression comprehension,'' in \emph{Proceedings of the IEEE/CVF Conference on Computer Vision and Pattern Recognition}, 2018.

\bibitem{wang2019neighbourhood}
P.~Wang, Q.~Wu, J.~Cao, C.~Shen, L.~Gao, and A.~v.~d. Hengel, ``Neighbourhood watch: Referring expression comprehension via language-guided graph attention networks,'' in \emph{Proceedings of the IEEE/CVF Conference on Computer Vision and Pattern Recognition}, 2019.

\bibitem{kamath2021mdetr}
A.~Kamath, M.~Singh, Y.~LeCun, G.~Synnaeve, I.~Misra, and N.~Carion, ``Mdetr-modulated detection for end-to-end multi-modal understanding,'' in \emph{Proceedings of the IEEE/CVF International Conference on Computer Vision}, 2021, pp. 1780--1790.

\bibitem{xiao2017weakly}
F.~Xiao, L.~Sigal, and Y.~Jae~Lee, ``Weakly-supervised visual grounding of phrases with linguistic structures,'' in \emph{Proceedings of the IEEE/CVF Conference on Computer Vision and Pattern Recognition}, 2017.

\bibitem{sun2021discriminative}
M.~Sun, J.~Xiao, E.~G. Lim, S.~Liu, and J.~Y. Goulermas, ``Discriminative triad matching and reconstruction for weakly referring expression grounding,'' \emph{IEEE transactions on pattern analysis and machine intelligence}, vol.~43, no.~11, pp. 4189--4195, 2021.

\bibitem{zhu2021utilizing}
H.~Zhu, A.~Sadhu, Z.~Zheng, and R.~Nevatia, ``Utilizing every image object for semi-supervised phrase grounding,'' in \emph{Proceedings of the IEEE/CVF Winter Conference on Applications of Computer Vision}, 2021, pp. 2210--2219.

\bibitem{chou2022semi}
S.-H. Chou, Z.~Fan, J.~J. Little, and L.~Sigal, ``Semi-supervised grounding alignment for multi-modal feature learning,'' in \emph{2022 19th Conference on Robots and Vision (CRV)}.\hskip 1em plus 0.5em minus 0.4em\relax IEEE, 2022, pp. 48--57.

\bibitem{subramanian2022reclip}
S.~Subramanian, W.~Merrill, T.~Darrell, M.~Gardner, S.~Singh, and A.~Rohrbach, ``Reclip: A strong zero-shot baseline for referring expression comprehension,'' in \emph{Proceedings of the 60th Annual Meeting of the Association for Computational Linguistics (Volume 1: Long Papers)}, 2022, pp. 5198--5215.

\bibitem{li2022adapting}
J.~Li, G.~Shakhnarovich, and R.~A. Yeh, ``Adapting clip for phrase localization without further training,'' \emph{arXiv preprint arXiv:2204.03647}, 2022.

\bibitem{lin2021m6}
J.~Lin, R.~Men, A.~Yang, C.~Zhou, Y.~Zhang, P.~Wang, J.~Zhou, J.~Tang, and H.~Yang, ``M6: Multi-modality-to-multi-modality multitask mega-transformer for unified pretraining,'' in \emph{Proceedings of the 27th ACM SIGKDD Conference on Knowledge Discovery \& Data Mining}, 2021, pp. 3251--3261.

\bibitem{li2021align}
J.~Li, R.~Selvaraju, A.~Gotmare, S.~Joty, C.~Xiong, and S.~C.~H. Hoi, ``Align before fuse: Vision and language representation learning with momentum distillation,'' \emph{Advances in neural information processing systems}, vol.~34, pp. 9694--9705, 2021.

\bibitem{wang2022unifying}
P.~Wang, A.~Yang, R.~Men, J.~Lin, S.~Bai, Z.~Li, J.~Ma, C.~Zhou, J.~Zhou, and H.~Yang, ``Unifying architectures, tasks, and modalities through a simple sequence-to-sequence learning framework,'' \emph{arXiv preprint arXiv:2202.03052}, 2022.

\bibitem{peng2023sgva}
F.~Peng, X.~Yang, L.~Xiao, Y.~Wang, and C.~Xu, ``Sgva-clip: Semantic-guided visual adapting of vision-language models for few-shot image classification,'' \emph{IEEE Transactions on Multimedia}, 2023.

\bibitem{bengio2009curriculum}
Y.~Bengio, J.~Louradour, R.~Collobert, and J.~Weston, ``Curriculum learning,'' in \emph{Proceedings of the 26th annual international conference on machine learning}, 2009, pp. 41--48.

\bibitem{shu2019transferable}
Y.~Shu, Z.~Cao, M.~Long, and J.~Wang, ``Transferable curriculum for weakly-supervised domain adaptation,'' in \emph{Proceedings of the AAAI Conference on Artificial Intelligence}, vol.~33, no.~01, 2019, pp. 4951--4958.

\bibitem{soviany2022curriculum}
P.~Soviany, R.~T. Ionescu, P.~Rota, and N.~Sebe, ``Curriculum learning: A survey,'' \emph{International Journal of Computer Vision}, pp. 1--40, 2022.

\bibitem{wang2021survey}
X.~Wang, Y.~Chen, and W.~Zhu, ``A survey on curriculum learning,'' \emph{IEEE Transactions on Pattern Analysis and Machine Intelligence}, 2021.

\bibitem{choi2019pseudo}
J.~Choi, M.~Jeong, T.~Kim, and C.~Kim, ``Pseudo-labeling curriculum for unsupervised domain adaptation,'' \emph{arXiv preprint arXiv:1908.00262}, 2019.

\bibitem{cascante2021curriculum}
P.~Cascante-Bonilla, F.~Tan, Y.~Qi, and V.~Ordonez, ``Curriculum labeling: Revisiting pseudo-labeling for semi-supervised learning,'' in \emph{Proceedings of the AAAI Conference on Artificial Intelligence}, vol.~35, no.~8, 2021, pp. 6912--6920.

\bibitem{zhang2021review}
L.~Zhang, Z.~Mao, B.~Xu, Q.~Wang, and Y.~Zhang, ``Review and arrange: Curriculum learning for natural language understanding,'' \emph{IEEE/ACM Transactions on Audio, Speech, and Language Processing}, vol.~29, pp. 3307--3320, 2021.

\bibitem{tay2019simple}
Y.~Tay, S.~Wang, A.~T. Luu, J.~Fu, M.~C. Phan, X.~Yuan, J.~Rao, S.~C. Hui, and A.~Zhang, ``Simple and effective curriculum pointer-generator networks for reading comprehension over long narratives,'' in \emph{Proceedings of the 57th Annual Meeting of the Association for Computational Linguistics}, 2019, pp. 4922--4931.

\bibitem{gong2016multi}
C.~Gong, D.~Tao, S.~J. Maybank, W.~Liu, G.~Kang, and J.~Yang, ``Multi-modal curriculum learning for semi-supervised image classification,'' \emph{IEEE Transactions on Image Processing}, vol.~25, no.~7, pp. 3249--3260, 2016.

\bibitem{zhao2022exploiting}
S.~Zhao, Z.~Zhang, S.~Schulter, L.~Zhao, A.~Stathopoulos, M.~Chandraker, D.~Metaxas \emph{et~al.}, ``Exploiting unlabeled data with vision and language models for object detection,'' \emph{arXiv preprint arXiv:2207.08954}, 2022.

\bibitem{kumar2010self}
M.~Kumar, B.~Packer, and D.~Koller, ``Self-paced learning for latent variable models,'' \emph{Advances in neural information processing systems}, vol.~23, 2010.

\bibitem{girshick2015fast}
R.~Girshick, ``Fast r-cnn,'' in \emph{Proceedings of the IEEE international conference on computer vision}, 2015, pp. 1440--1448.

\bibitem{rezatofighi2019generalized}
H.~Rezatofighi, N.~Tsoi, J.~Gwak, A.~Sadeghian, I.~Reid, and S.~Savarese, ``Generalized intersection over union: A metric and a loss for bounding box regression,'' in \emph{Proceedings of the IEEE/CVF conference on computer vision and pattern recognition}, 2019, pp. 658--666.

\bibitem{cong2022reltr}
Y.~Cong, M.~Y. Yang, and B.~Rosenhahn, ``Reltr: Relation transformer for scene graph generation,'' \emph{arXiv preprint arXiv:2201.11460}, 2022.

\bibitem{cornia2020meshed}
M.~Cornia, M.~Stefanini, L.~Baraldi, and R.~Cucchiara, ``Meshed-memory transformer for image captioning,'' in \emph{Proceedings of the IEEE/CVF conference on computer vision and pattern recognition}, 2020, pp. 10\,578--10\,587.

\bibitem{mokady2021clipcap}
R.~Mokady, A.~Hertz, and A.~H. Bermano, ``Clipcap: Clip prefix for image captioning,'' \emph{arXiv preprint arXiv:2111.09734}, 2021.

\bibitem{yao2021cpt}
Y.~Yao, A.~Zhang, Z.~Zhang, Z.~Liu, T.-S. Chua, and M.~Sun, ``Cpt: Colorful prompt tuning for pre-trained vision-language models,'' \emph{arXiv preprint arXiv:2109.11797}, 2021.

\bibitem{liu2019adaptive}
X.~Liu, L.~Li, S.~Wang, Z.-J. Zha, D.~Meng, and Q.~Huang, ``Adaptive reconstruction network for weakly supervised referring expression grounding,'' in \emph{Proceedings of the IEEE/CVF International Conference on Computer Vision}, 2019.

\bibitem{liu2019knowledge}
X.~Liu, L.~Li, S.~Wang, Z.-J. Zha, L.~Su, and Q.~Huang, ``Knowledge-guided pairwise reconstruction network for weakly supervised referring expression grounding,'' in \emph{Proceedings of the 27th ACM International Conference on Multimedia}, 2019, pp. 539--547.

\bibitem{li2021referring}
M.~Li and L.~Sigal, ``Referring transformer: A one-step approach to multi-task visual grounding,'' \emph{Advances in Neural Information Processing Systems}, vol.~34, pp. 19\,652--19\,664, 2021.

\bibitem{du2022visual}
Y.~Du, Z.~Fu, Q.~Liu, and Y.~Wang, ``Visual grounding with transformers,'' in \emph{2022 IEEE International Conference on Multimedia and Expo (ICME)}.\hskip 1em plus 0.5em minus 0.4em\relax IEEE, 2022, pp. 1--6.

\end{thebibliography}

%%%%%%%%%%%%%%%%%%%%%%%%%%%%%%%%%%%%%%%%%%%%%%%%%%%%%%%%%%%%%%%%%%%%%%%%%%%
%----------------------
%----------------------
%----------------------
% \newpage
% \input{bios/bios}

\vfill
\end{document}